\documentclass[sigconf]{aamas} 

\usepackage{balance} 
\usepackage{enumitem}
\usepackage{subcaption}

\setcopyright{ifaamas}
\acmConference[AAMAS '23]{Proc.\@ of the 22nd International Conference
on Autonomous Agents and Multiagent Systems (AAMAS 2023)}{May 29 -- June 2, 2023}
{London, United Kingdom}{A.~Ricci, W.~Yeoh, N.~Agmon, B.~An (eds.)}
\copyrightyear{2023}
\acmYear{2023}
\acmDOI{}
\acmPrice{}
\acmISBN{}

\acmSubmissionID{1230}

\title[]{Towards Computationally Efficient Responsibility Attribution in Decentralized Partially Observable MDPs}

\author{Stelios Triantafyllou}
\affiliation{
  \institution{Max Planck Institute for Software Systems}
  \city{Saarbrücken}
  \country{Germany}}
\email{strianta@mpi-sws.org}

\author{Goran Radanovic}
\affiliation{
  \institution{Max Planck Institute for Software Systems}
  \city{Saarbrücken}
  \country{Germany}}
\email{gradanovic@mpi-sws.org}

\begin{abstract}

Responsibility attribution is a key concept of accountable multi-agent decision making. Given a sequence of actions, responsibility attribution mechanisms quantify the impact of each participating agent to the final outcome. One such popular mechanism is based on actual causality, and it assigns (causal) responsibility based on the actions that were found to be pivotal for the considered outcome. However, the inherent problem of pinpointing actual causes and consequently determining the exact responsibility assignment has shown to be computationally intractable. In this paper, we aim to provide a practical algorithmic solution to the problem of responsibility attribution under a computational budget. We first formalize the problem in the framework of Decentralized Partially Observable Markov Decision Processes (Dec-POMDPs) augmented by a specific class of Structural Causal Models (SCMs). Under this framework, we introduce a Monte Carlo Tree Search (MCTS) type of method which efficiently approximates the agents' degrees of responsibility. This method utilizes the structure of a novel search tree and a pruning technique, both tailored to the problem of responsibility attribution. Other novel components of our method are (a) a \textit{child selection policy} based on \textit{linear scalarization} and (b) a \textit{backpropagation procedure} that accounts for a minimality condition that is typically used to define actual causality. We experimentally evaluate the efficacy of our algorithm through a simulation-based test-bed, which includes three team-based card games.

\end{abstract}


\newcommand{\BibTeX}{\rm B\kern-.05em{\sc i\kern-.025em b}\kern-.08em\TeX}

\newcommand{\agentzero}{{$ \text{Ag}0$}}
\newcommand{\agentone}{{$ \text{Ag}1$}}

\newcommand{\causalsetting}{{$ (\mathcal{C}, \vec{u})$}}
\newcommand{\witness}{$ (\vec{W}, \vec{w}', \vec{a}')$}
\newcommand{\actualcause}{$ \vec{A} = \vec{a}$}
\newcommand{\acwpair}{$ (\vec{A} = \vec{a}, (\vec{W}, \vec{w}', \vec{a}'))$}

\DeclareMathOperator*{\argmax}{arg\,max}

\begin{document}


\pagestyle{fancy}
\fancyhead{}


\maketitle 


\newtoggle{arxiv}
\settoggle{arxiv}{true}


\section{Introduction}\label{sec.intro}

One of the well known {\em Gedankenexperimente} in the AI literature on actual causality and responsibility attribution is the story of \textit{Suzy and Billy}. As J. Y. Halpern describes it in his book on {\em Actual Causality} \cite{halpern2016actual}, the story goes as follows: 

``{\em Suzy and Billy both pick up rocks and throw them at a bottle. Suzy’s rock gets there first, shattering the bottle. Because both throws are perfectly accurate, Billy’s would have shattered the bottle had it not been preempted by Suzy’s throw.}''

Who is to be held responsible for the bottle being shattered? As the curious reader may have noticed, the conceptual challenge of attributing responsibility for this outcome lies in the fact that the outcome would not have changed had Suzy not thrown her rock. Needless to say, there are a plethora of other examples from moral philosophy that challenge human intuition on how responsibility should be ascribed, including {\em Bogus Prevention}~\cite{hiddleston2005causal}, {\em Marksmen}~\cite{halpern2016actual}, {\em Arsonists}~\cite{halpern2005causes}, and {\em Bystanders}~\cite{halpern2011actual}. 

Much of the (recent) work in moral philosophy and AI has focused on resolving these conceptual challenges utilizing different formal frameworks \cite{baier2021game, baier2021responsibility, alechina2020causality, yazdanpanah2019strategic}.
Often, perhaps unsurprisingly, these works \cite{chockler2004responsibility, halpern2018towards, 10.1145/3514094.3534133} take as a starting point the framework of actual causality based on Structural Causal Models (SCMs) \cite{pearl1995causal}. 
Given a specific scenario, approaches that utilize this framework typically first pinpoint actions that were pivotal to the outcome of that scenario.
It is also not hard to see the importance of these works for accountable AI systems. 
Consider, for example, a semi-autonomous vehicle, and let Suzy be the auto-pilot of this vehicle, Billy be the human driver that oversees the autopilot, and the shattered bottle be the pedestrian injured in an accident caused by the vehicle. 

However, real-world scenarios are often much more complex than the aforementioned examples from moral philosophy capture. In order to operationalize responsibility attribution for automated, yet accountable decision making systems it is important to ground it in a framework that is general enough to capture the nuances of real-world decision making settings. This has recently been recognized by \citet{10.1145/3514094.3534133}, who study the problem of actual causality and responsibility attribution in Decentralized Partially Observable Markov Decision Processes (Dec-POMDPs) -- a rather general framework for multi-agent sequential decision making under uncertainty.
However, while \citet{10.1145/3514094.3534133} show how to combine Dec-POMDPs with SCMs to enable causal reasoning, they still primarily focus on challenges related to defining actual causality and designing responsibility attribution mechanisms.
In contrast, in this paper we focus on the 
fact that the problem of inferring actual causes, and consequently attributing causal responsibility, is known to have high computational complexity~\cite{eiter2002complexity, aleksandrowicz2017computational, halpern2015modification}.
Since determining the exact responsibility assignment is then intractable, we ask the following question:
``{\em Can we design an efficient procedure for approximately ascribing responsibility in Dec-POMDPs?}''

\begin{figure*}[ht]
	\centering
	\begin{subfigure}{0.99\textwidth}
    	\centering
		\resizebox{0.99\linewidth}{!}{\includegraphics{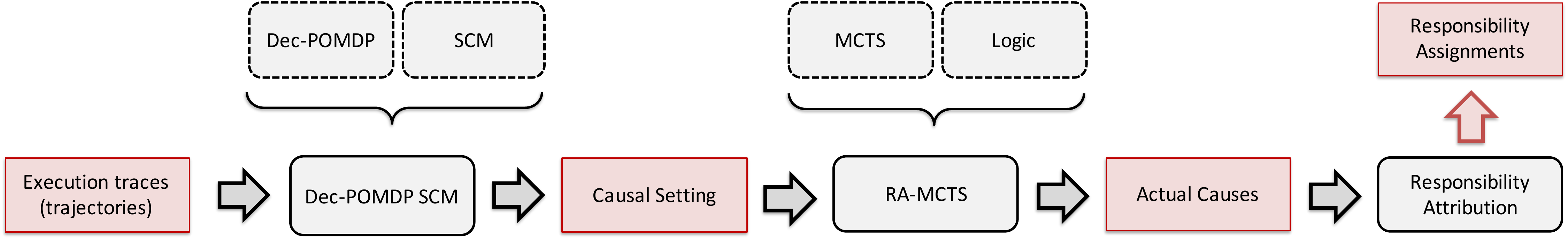}}
	\end{subfigure}
    \caption{The figure provides an overview of our approach to responsibility attribution in Dec-POMDPs. Based on received execution traces, that is the agents' trajectories, our approach first aims to infer the underlying context under which these decisions were made. To do so we utilize Dec-POMDP SCMs -- a framework introduced in \cite{10.1145/3514094.3534133} that combines Dec-POMDPs and SCMs. The context in our case models randomness/noise, and together with the underlying causal model forms the causal setting. 
    The next step is  to apply an MCTS type of search to infer actual causes that are consistent with the definition of actual causality introduced in \cite{10.1145/3514094.3534133}.
    This search method is formally specified in a language that is an extension of propositional logic. 
    In order to determine the agents' degrees of responsibilities, we apply the responsibility attribution method of \cite{chockler2004responsibility} over the actual causes found by the search.
    In cases where the underlying context cannot be exactly inferred we use posterior inference over possible contexts. More specifically, we first draw samples from the posterior over possible contexts, and then repeat the process described above for each sampled context. Agents are assigned the average degrees of responsibility over all samples.
    }
	\label{fig.intro}
\end{figure*}

We propose an algorithmic framework for approximating  responsibility assignments under a computational budget. Having a bounded budget is important for complex systems, where brute-force approaches do not work \cite{halpern2016actual, ibrahim2019efficient}.
The example scenarios range from autonomous public transit systems and system controllers, such as traffic light controllers (TLC), to multi-agent cooperative systems, such as warehouse robots. For an extended discussion on the application scenario of TLC see \iftoggle{arxiv}{Appendix \ref{app.tlc}}{the Supplemental Material}.

Fig. \ref{fig.intro} provides an overview of our algorithmic approach. We recognize two main challenges that this approach has to overcome. The first one is a statistical challenge and is related to the fact that in practice the {\em context} under which an outcome of interest is generated cannot be always inferred. As explained in Fig. \ref{fig.intro}, to make responsibility attribution feasible in such cases, we use posterior inference over the possible values of the underlying context.
The second challenge is a computational one and it is related to the computational complexity of identifying actual causes. We tackle this challenge by applying a Monte Carlo Tree Search (MCTS) type of method tailored to the problem of finding actual causes. 
As we show in this paper, our approach significantly outperforms baselines in terms of approximating the “ideal” responsibility assignments, obtained under no uncertainty and unlimited computational budget. 
Our contributions are primarily related to the design and experimental evaluation of this algorithmic framework, and they include:
\begin{itemize}[leftmargin=*]
    \item A novel {\bf search tree} tailored to the tasks of pinpointing actual causes and attributing responsibility. 
    \item A novel {\bf pruning technique} that 
    utilizes the structural properties of both the actual causality definition of \citet{10.1145/3514094.3534133} and the responsibility attribution mechanism of \citet{chockler2004responsibility}.
    \item {\bf Responsibility Attribution-MCTS (RA-MCTS)}, a new {\bf search method} for efficiently finding actual causes under a given \textit{causal setting}. 
    Compared to standard MCTS, the main novel components of RA-MCTS are in its \textit{simulation phase}, \textit{evaluation function}, \textit{child selection policy}, and \textit{backpropagation phase}.
    \item {\bf Experimental test-bed} for evaluating the efficacy of RA-MCTS.
    The test-bed is based on three card games, \textit{Euchre}~\cite{wright2001mathematics}, \textit{Spades}~\cite{cohensius2019bidding}, and a team variation of the game \textit{Goofspiel}~\cite{ross1971goofspiel}.
    We deem the test-bed to be generally useful for studying actual causality and responsibility attribution in multi-agent sequential decision making. 
    Our experimental results show that RA-MCTS almost always outperforms baselines, such as random search, brute-force search, or modifications of RA-MCTS.
    Our results also show that in cases where the underlying context cannot be exactly inferred, computing a good approximation of the ``ideal'' responsibility assignment might not be possible, even under unlimited computational budget.
    This can happen when the posterior distribution over the possible contexts is not informative enough.
    \footnote{Code to reproduce the experiments is available at \href{https://github.com/stelios30/aamas23-responsibility-attribution-mcts.git}{https://github.com/stelios30/aamas23-responsibility-attribution-mcts.git}.}
\end{itemize}


\subsection{Additional Related Work}

This paper is related to works on responsibility and blame attribution in multi-agent decision making \cite{baier2021game, baier2021responsibility,  yazdanpanah2019strategic, triantafyllou2021blame, halpern2018towards, friedenberg2019blameworthiness}.

To the best of our knowledge, there is no prior work on developing general algorithmic approaches on efficiently computing degrees of causal responsibility.
The closest we could find, are domain-specific applications of the Chockler and Halpern responsibility approach \cite{chockler2004responsibility} in program verification \cite{chockler2008causes, chockler2008efficient}. 
Chapter $8$ of \cite{halpern2016actual} provides a general overview of such applications.
Additionally, to our knowledge, the only general algorithmic approach on determining causality, and subsequently responsibility attribution, is that of \cite{ibrahim2019efficient}. Their approach on checking actual causality utilizes SAT solvers and thus is significantly different than ours. They also restrict their focus to binary models, as opposed to ours which considers categorical variables.

The only other work that has used the same framework as the one used in this paper is that of \citet{10.1145/3514094.3534133}. 
Close to our work in this aspect, \citet{buesing2018woulda} and \citet{oberst2019counterfactual} have considered a combination of SCMs with POMDPs. \citet{tsirtsis2021counterfactual} utilize a connection between SCMs and MDPs to generate counterfactual explanations for 
sequential decision making.

This paper is also related to a line of work which introduces variants of MCTS that apply to specific domains. For instance, \citet{schadd2008single} and \citet{bjornsson2009cadiaplayer} propose modifications to MCTS in order to adapt it to single-player games. We refer the interested reader to \citet{browne2012survey} for more such examples.


\section{Framework and Background}\label{sec.setting}

In this section, we first give an overview of a formal framework which allows us to study responsibility attribution in the context of multi-agent sequential decision making. This framework is adopted from \citet{10.1145/3514094.3534133}, and relies on decentralized partially observable Markov decision processes (Dec-POMDPs) \cite{bernstein2002complexity, oliehoek2016concise} and structural causal models (SCMs) \cite{pearl2009causality, peters2017elements}.
Next, we provide the necessary background on actual causality and responsibility attribution.
Finally, we state the responsibility attribution problem and highlight its main algorithmic challenges.

\subsection{Dec-POMDPs}\label{sec.decpomdp}

The first component of this framework are Dec-POMDPs with $n$ agents; state space $\mathcal{S}$; joint action space $\mathcal{A} = \times_{i = 1}^n \mathcal{A}_i$, where $\mathcal{A}_i$ is the action space of agent $i$; transition probability function $P: \mathcal{S} \times \mathcal{A} \rightarrow \Delta(\mathcal{S})$; joint observation space $\mathcal{O} = \times_{i = 1}^n \mathcal{O}_i$, where $\mathcal{O}_i$ is the observation space of agent $i$; observation probability function $\Omega: \mathcal{S} \rightarrow \Delta(\mathcal{O})$; finite time horizon $T$; initial state distribution $\sigma$. Here $\Delta$ denotes the probability simplex.
For ease of notation, rewards are considered to be part of observations.

Each agent $i$ is modeled with an information state space $\mathcal{I}_i$; decision making policy $\pi_i: \mathcal{I}_i \rightarrow \Delta(\mathcal{A}_i)$; information probability function $Z_i: \mathcal{I}_i \times \mathcal{A}_i \times \mathcal{O}_i \rightarrow \mathcal{I}_i$; initial information probability function $Z_{i,0}: \mathcal{O}_i \rightarrow \mathcal{I}_i$.
We denote with $\pi_i(a_{i}|\imath_i)$ agent $i$'s probability of taking action $a_i$ given information state $\imath_i$, and with $\pi$ the collection of all agents' policies, i.e., the agents' joint policy.

We assume spaces $\mathcal{S}$, $\mathcal{A}$, $\mathcal{O}$ and $\mathcal{I}_i$ to be finite and discrete.

\subsection{Dec-POMDPs and Structural Causal Models}\label{sec.decpomdp_scm}

In order to reason about actual causality and responsibility attribution in multi-agent sequential decision making \citet{10.1145/3514094.3534133} view Dec-POMDPs as SCMs.\footnote{They establish a connection between the two by building on prior work from \citet{buesing2018woulda}.} More specifically, given a Dec-POMDP $\mathcal{M} = (\mathcal{S}, \{1, ..., n\}, \mathcal{A}, P, \mathcal{O},  \Omega, T, \sigma)$ and a model $m_i = (\mathcal{I}_i, \pi_i, Z_i, Z_{i,0})$ for each agent $i$, they construct a SCM $\mathcal{C}$, which they refer to as Dec-POMDP SCM. Under $\mathcal{C}$ functions $P$, $\Omega$, $\{Z_i\}_{i\in\{1,...,n\}}$ and $\{\pi_i\}_{i \in \{1, .., n\}}$ are parameterized as follows
\begin{align}\label{eq.struct_eq}
    &S_t = g_{S_t}(S_{t-1}, A_{t-1}, U_{S_t}), 
    \quad O_t = g_{O_t}(S_{t}, U_{O_t}),\nonumber\\
    &I_{i,t} = g_{I_{i, t}}(I_{i,t-1}, A_{i,t-1}, O_{i,t}, U_{I_{i,t}}), 
    \quad A_{i,t} = g_{A_{i, t}}(I_{i,t}, U_{A_{i,t}}),
\end{align}
where $g_{S_t}$, $g_{O_t}$, $g_{I_{i, t}}$ and $g_{A_{i, t}}$ are deterministic functions, and $U_{S_t}$, $U_{O_t}$, $U_{I_{i,t}}$ and $U_{A_{i,t}}$ are independent noise variables with dimensions $|\mathcal{S}|$, $|\mathcal{O}|$, $|\mathcal{I}_i|$ and $|\mathcal{A}_i|$, respectively.\footnote{Such a parameterization is always possible \cite{10.1145/3514094.3534133}.}

Following SCM terminology \cite{pearl2009causality}, we refer to state variables $S_t$, observation variables $O_t$, information variables $I_{i,t}$ and action variables $A_{i,t}$ as the endogenous variables of $\mathcal{C}$. Furthermore, we call noise variables $U$ the exogenous variables of $\mathcal{C}$ and a setting $\vec{u}$ of $U$ context.
Note that given a context $\vec{u}$ one can compute the value of any endogenous variable in $\mathcal{C}$ by consecutively solving equations in \eqref{eq.struct_eq}, also called structural equations. 
Therefore, a Dec-POMDP SCM-context pair \causalsetting, also called causal setting, specifies a unique trajectory $\tau = \{(s_t, a_t)\}_{t=0}^{T-1}$.

Another well-known notion in causality that is important for our analysis is that of interventions \cite{pearl1995causal}.\footnote{In this paper, we consider interventions on action variables only.} An intervention $A_{i,t} \leftarrow a'_i$ on SCM $\mathcal{C}$ is performed by replacing $g_{A_{i, t}}(I_{i,t}, U_{A_i,t})$ in Eq. \eqref{eq.struct_eq} with $a'_i$, also called the counterfactual action of the intervention.
We denote the resulting SCM by $\mathcal{C}^{A_{i,t} \leftarrow a'_i}$.
If one has knowledge over $\mathcal{C}$ as well as the context $\vec{u}$ under which a trajectory $\tau$ was generated, they can efficiently compute the counterfactual outcome of that trajectory under some intervention $A_{i,t} \leftarrow a'_i$ on $\mathcal{C}$. This can be done by simply generating the counterfactual trajectory $\tau^{cf}$ that corresponds to the causal setting $(\mathcal{C}^{A_{i,t} \leftarrow a'_i}, \vec{u})$. In other words, they can predict exactly what would have happened in that scenario had agent $i$ taken action $a'_i$ instead of $a_{i, t}$. 
However, the true underlying SCM $\mathcal{C}$ or context $\vec{u}$ are not always available in practice.
Following a standard modeling approach \cite{10.1145/3514094.3534133, lorberbom2021learning, tsirtsis2021counterfactual}, we restrict our focus on a specific class of SCMs, the Gumbel-Max SCMs, introduced by \citet{oberst2019counterfactual}.
More details on Gumbel-Max SCMs and how they can be integrated in the Dec-POMDP SCM framework, can be found in \iftoggle{arxiv}{Appendix \ref{app.gumbel}}{the Supplemental Material}.\footnote{For a more detailed overview of SCMs we refer the reader to \cite{pearl2009causality}.}

\subsection{Actual Causality}\label{sec.ac}

Next, we present a language for reasoning about actual causality in (Dec-POMDP) SCMs \cite{halpern2016actual}. Let $\mathcal{C}$ be a Dec-POMDP SCM. A primitive event in $C$ is any formula of the form $V = v$, where $V$ is an endogenous variable of $\mathcal{C}$ and $v$ is a valid value of $V$. We say that a Boolean combination of primitive events constitutes an event.
Given a context $\vec{u}$ and an event $\phi$, we write $(\mathcal{C}, \vec{u}) \models \phi$ to denote that $\phi$ takes place in the causal setting \causalsetting. Furthermore, for a set of interventions $\vec{A} \leftarrow \vec{a}'$ on $\mathcal{C}$, we write $(\mathcal{C}, \vec{u}) \models [\vec{A} \leftarrow \vec{a}']\phi$, if $(\mathcal{C}^{\vec{A} \leftarrow \vec{a}'}, \vec{u}) \models \phi$.
For example, let $\tau = \{(s_t, a_t)\}_{t=0}^{T-1}$ be the trajectory that corresponds to \causalsetting. Consider the counterfactual scenario in which agent $i$ takes action $a_i'$ instead of $a_{i, t}$ in $\tau$, and the process transitions to state $s$ at $t+1$. This can be expressed by
\begin{align*}
    (\mathcal{C}, \vec{u}) \models [A_{i,t} \leftarrow a_i'](S_{t+1} = s).
\end{align*}

In the context of Dec-POMDP SCMs actual causality is related to the process of pinpointing agents' actions that were critical for $\phi$ to happen in \causalsetting. In this paper, we adopt the actual cause definition proposed by \citet{10.1145/3514094.3534133}. Their definition utilizes the agents' information states in order to explicitly account for the temporal dependencies between agents' actions.

\begin{definition}\label{def.ac}
(Actual Cause)
\actualcause~is an actual cause of the event $\phi$ in \causalsetting under the contingency $\vec{W} = \vec{w}'$ if the following conditions hold:
\begin{enumerate}[label=AC\arabic*.]
    \item $(\mathcal{C}, \vec{u}) \models (\vec{A} = \vec{a})$ and $(\mathcal{C}, \vec{u}) \models \phi$
    \item There is a setting $\vec{a}'$ of the variables in $\vec{A}$, such that
    \begin{align*}
        (\mathcal{C}, \vec{u}) \models [\vec{A} \leftarrow \vec{a}', \vec{W} \leftarrow \vec{w}']\neg\phi
    \end{align*}
    \item $\vec{A} \cup \vec{W}$ is minimal w.r.t. conditions \textit{AC1} and \textit{AC2}
    \item For every agent $i$ and time-step $t$ such that $A_{i,t} \in \vec{A}$ and $(\mathcal{C}, \vec{u}) \models (I_{i,t} = \imath_{i,t})$, it holds that
    \begin{align*}
        (\mathcal{C}, \vec{u}) \models [\vec{A} \leftarrow \vec{a}', \vec{W} \leftarrow \vec{w}'] (I_{i,t} = \imath_{i, t})
    \end{align*}
    \item For every agent $i$ and time-step $t$ such that $A_{i,t} \in \vec{W}$ and $(\mathcal{C}, \vec{u}) \models (I_{i,t} = \imath_{i,t})$, it holds that
    \begin{align*}
        (\mathcal{C}, \vec{u}) \models [\vec{A} \leftarrow \vec{a}', \vec{W} \leftarrow \vec{w}'] \neg (I_{i,t} = \imath_{i,t})
    \end{align*}
\end{enumerate}
We say that the tuple \witness~is a witness of \actualcause~being an actual cause of $\phi$ in \causalsetting.
\end{definition}

\textit{AC1} requires that both \actualcause~and $\phi$ happened in \causalsetting.
\textit{AC2} implies that $\phi$ would not have occurred under the interventions $\vec{A} \leftarrow \vec{a}'$ and $\vec{W} \leftarrow \vec{w}'$ on \causalsetting.
\textit{AC3} is a minimality condition, which ensures that there are no subsets $\vec{A}'$ and $\vec{W}'$ of $\vec{A}$ and $\vec{W}$, and setting $\vec{w}''$ of $\vec{W}'$, such that $\vec{A}' = \vec{a}'$ and $\vec{W}' = \vec{w}''$ satisfy \textit{AC1} and \textit{AC2}, where $\vec{a}'$ is the restriction of $\vec{a}$ to the variables of $\vec{A}$.
\textit{AC4} (resp. \textit{AC5}) requires that the information states which correspond to the action variables in $\vec{A}$ (resp. $\vec{W}$) have the same (resp. different) values in the counterfactual scenario $(\mathcal{C}^{\vec{A} \leftarrow \vec{a}', \vec{W} \leftarrow \vec{w}'}, \vec{u})$ and the actual scenario \causalsetting.
We say that a conjunct of an actual cause \actualcause~constitutes a part of that cause. 
If for some \actualcause~and $\vec{W} = \vec{w}'$ conditions \textit{AC1}, \textit{AC2}, \textit{AC4} and \textit{AC5} hold we say that \actualcause~is a candidate actual cause of $\phi$ in \causalsetting~under the contingency $\vec{W} = \vec{w}'$.
We also say that a set of interventions $\vec{X} \leftarrow \vec{x}'$ constitutes an (candidate) actual cause-witness pair according to Definition \ref{def.ac} if there exists such a pair \acwpair, where $\vec{X} = \vec{A} \cup \vec{W}$, and $\vec{a}'$ and $\vec{w}'$ are the projections of $\vec{x}'$ in $\vec{A}$ and $\vec{W}$, respectively.
 
\subsection{Responsibility Attribution}\label{sec.resp}

Responsibility attribution is a concept closely related to actual causality, which aims to determine the extent to which agents' actions were pivotal for some outcome.
In this paper, we adopt a responsibility attribution approach which was first introduced by \citet{chockler2004responsibility}, and then adapted by \citet{10.1145/3514094.3534133} to the setting of Dec-POMDP SCMs. Given a causal setting \causalsetting~and an event $\phi$, the Chockler and Halpern approach (henceforth CH) uses the following function to determine an agent $i$'s degree of responsibility for $\phi$ in \causalsetting~relative to a set of interventions $\vec{X} \leftarrow \vec{x}'$ on $\mathcal{C}$ and an actual causality definition $\mathcal{D}$
\begin{align}\label{eq.ch}
    dr_i((\mathcal{C}, \vec{u}), \phi, \vec{X} \leftarrow \vec{x}', \mathcal{D}) =
    \frac{m_i}{|\vec{X}|},
\end{align}
where $m_i$ is computed as follows. If $\vec{X} \leftarrow \vec{x}'$ constitutes an actual cause-witness pair \acwpair~of $\phi$ in \causalsetting~according to $\mathcal{D}$, then $m_i$ denotes the number of $i$'s action variables in $\vec{A}$. Otherwise, $m_i$ is $0$.
In this paper, an agent's degree of responsibility according to the CH approach is computed as follows.

\begin{definition}\label{def.ch}
(CH)
Consider a causal setting \causalsetting~and an event $\phi$ such that $(\mathcal{C}, \vec{u}) \models \phi$. With $\mathcal{D}$ being Definition \ref{def.ac}, an agent $i$'s degree of responsibility for $\phi$ in \causalsetting~is equal to the maximum value $dr_i((\mathcal{C}, \vec{u}), \phi, \vec{X} \leftarrow \vec{x}', \mathcal{D})$ over all possible sets of interventions $\vec{X} \leftarrow \vec{x}'$ on $\mathcal{C}$.
\end{definition}

The CH definition captures some key ideas of responsibility attribution. First, an agent's degree of responsibility depends on the size of an actual cause \actualcause~the agent participates in. Second, it depends on the amount of participation the agent has in that cause. Finally, it depends on the size of the smallest contingency of \actualcause, i.e., the minimum number of interventions that need to be performed on $\mathcal{C}$ in order to make $\phi$ counterfactually depend on $\vec{A}$.

\subsection{Problem Statement and Challenges}\label{sec.prob_chall}

Given a trajectory $\tau$ generated by causal setting \causalsetting, the general problem we are interested in is computing the agents' degrees of responsibility for the final outcome $\phi^\tau$ of $\tau$.
In this paper, we focus on two main challenges of this problem. The first one is related to the computational complexity of the problem. The second one has to do with the fact that in practice context $\vec{u}$ might not be known.

In order to address the first challenge, we view responsibility attribution as a \textbf{multi-objective search problem} with limited computational resources.
An algorithmic solution to this problem should find a set of interventions, for each agent, that maximizes the function in Eq. \eqref{eq.ch}.
The pipeline we consider for such algorithms can be summarized as follows. First, the algorithm searches for sets of interventions that constitute actual cause-witness pairs of the outcome $\phi^\tau$. Next, based on the found actual cause-witness pairs the algorithm computes the responsibility assignment.
A natural question that arises is how to choose which intervention sets to evaluate before the computational budget is exhausted. We believe that the answer to this question lies in the structural properties of Definitions \ref{def.ac} and \ref{def.ch} (Sections \ref{sec.tree}-\ref{sec.mcts}).
Another question related to this problem is how to recognize if a set of interventions is in fact an actual cause-witness pair. Even though it is easy to infer whether a set of interventions constitutes a candidate actual cause-witness pair of $\phi^\tau$ when \causalsetting~is known, it is impossible to know if it is minimal, i.e., if it satisfies condition \textit{AC3}, unless all of its subsets are first checked for \textit{AC1} and \textit{AC2}. 
Despite that, there are countermeasures that one can implement to reduce the negative impact that \textit{AC3} might have on the search process (Section \ref{sec.mcts}).

To address the second challenge, we view responsibility attribution as an \textbf{inference problem}. Our approach is to build on the above mentioned search algorithm, and by using posterior inference design a mechanism that can efficiently estimate responsibility assignments under context uncertainty (Section \ref{sec.alg_unc}).


\section{Algorithmic Solution}\label{sec.algorithm}

In this section, we analyze our algorithmic solutions to the search and inference problems described in Section \ref{sec.prob_chall}.
First, we propose a novel search tree tailored to the tasks of pinpointing actual causes and attributing responsibility. Next, we propose a pruning technique that utilizes the structural properties of Definitions \ref{def.ac} and \ref{def.ch}.
We then propose RA-MCTS, a novel Monte Carlo Tree Search (MCTS) type of method for finding approximate responsibility assignments under limited computational budget. 
Finally, we propose an extension of RA-MCTS to the unknown context regime.

\subsection{Search Tree}\label{sec.tree}

\begin{figure}[t]
    \centering
    \includegraphics[width=.75\columnwidth]{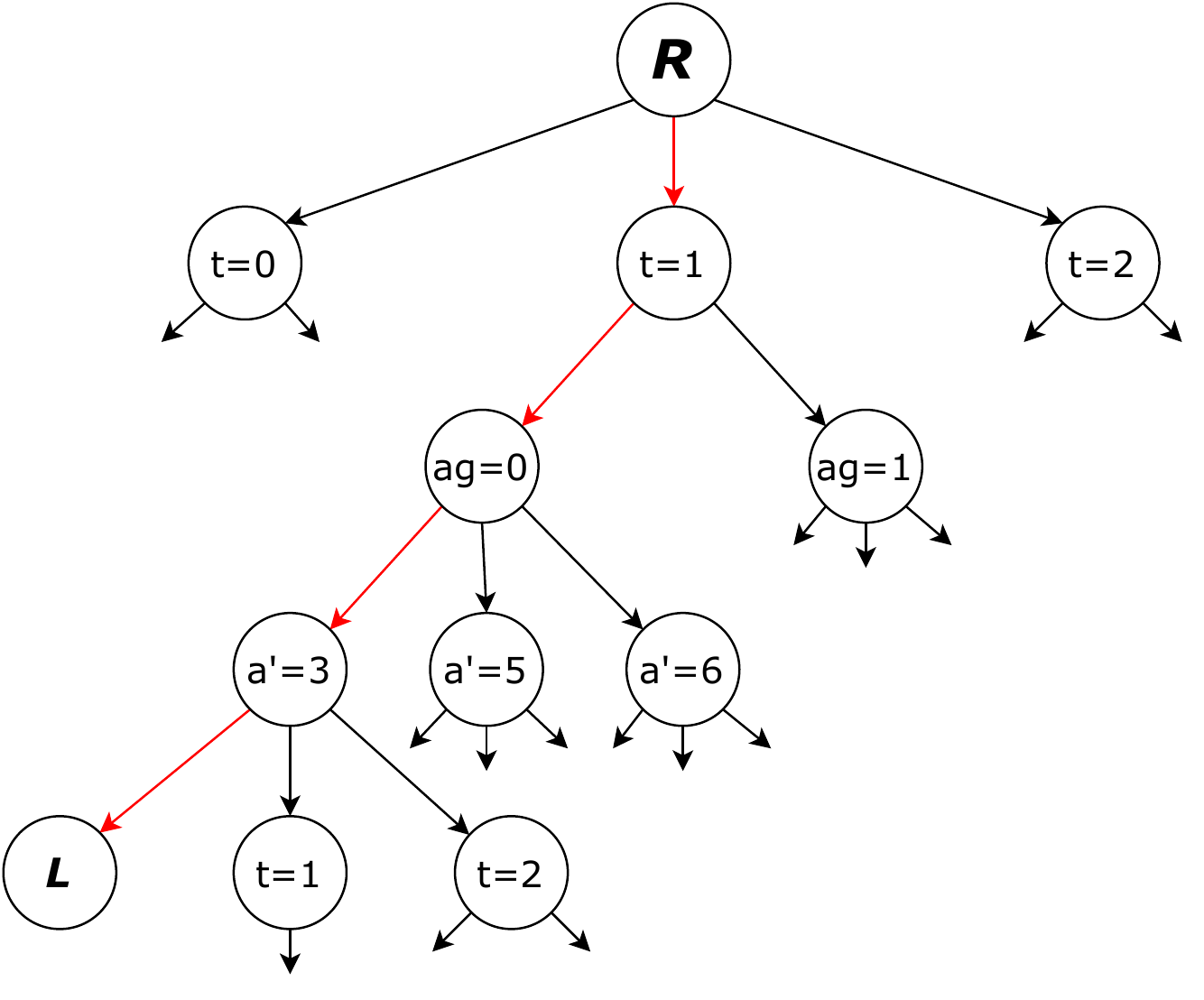}
    \caption{The red edges denote the path in our search tree, which corresponds to the intervention on the action of agent $0$ at time-step $1$ with counterfactual action $a'=3$, $A_{0,1} \leftarrow 3$.}
    \label{fig:search_tree}
\end{figure}
Fig. \ref{fig:search_tree} illustrates an instantiation of our proposed search tree. Note that the tree is defined relative to the causal setting \causalsetting, that is, every state, observation, information state and action is deterministically computed by the structural equations of $\mathcal{C}$ together with context $\vec{u}$ (see Eq. \eqref{eq.struct_eq}). Nodes in this tree fall into one of $5$ categories.
At the top of the tree, we have the \textit{Root} node, where the time-step of the first intervention is selected. Nodes $t=0$, $t=1$ and $t=2$ in Fig. \ref{fig:search_tree} correspond to their respective time-steps, and we call them \textit{TimeStep} nodes. From a \textit{TimeStep} node, the agent of the next intervention is picked.
Nodes $ag=0$ and $ag=1$ correspond to agents $0$ and $1$, and they are categorized as \textit{Agent} nodes. From an \textit{Agent} node, the counterfactual action of the next intervention is chosen from the available options.
More specifically, let $ag=i$ be the node where the search is currently on, and $t = t'$ be that node's parent. Let also $\vec{X} \leftarrow \vec{x}'$ denote the current set of interventions encoded in $ag=i$.
The available options from node $ag=i$ then include all the valid actions that $i$ could have taken at time-step $t'$, except the action that it would have normally taken given the current set of interventions, i.e., the action determined by the causal setting $(\mathcal{C}^{\vec{X} \leftarrow \vec{x}'}, \vec{u})$. 
Nodes $a'=3$, $a'=5$ and $a'=6$ in Fig. \ref{fig:search_tree} correspond to such counterfactual actions and they are characterized as \textit{Action} nodes. From an \textit{Action} node, search can either stop growing the intervention set, and hence transition to a \textit{Leaf} node $L$ or continue by transitioning to the next \textit{TimeStep} node.
If search transitions to $L$, then the current set of interventions is evaluated.
In case this set of interventions is found to change the final outcome $\phi^\tau$, it is added to the set of found candidate actual cause-witness pairs.

\subsection{Pruning}\label{sec.pruning}
Apart from its intuitive nature and computational efficiency (Section \ref{sec.results}), the search tree of Fig. \ref{fig:search_tree} also allows us to apply a number of effective pruning techniques. Pruning can take place at any point during the search and it is basically the process of removing branches from a tree that cannot possibly improve the output of the algorithm. In our setting, this means that a node needs not be further visited if it becomes apparent that the evaluation of any leaf node reachable from that node cannot in any way influence the final responsibility assignment. Our method prunes away a node (and all of its descendants) if any of the following conditions hold:
\begin{itemize}[leftmargin=*]
    \item It is a \textit{Leaf} node that has already been evaluated.
    \item It is the closest ancestor \textit{Agent} node of a \textit{Leaf} node $L$, such that $L$'s encoded set of interventions $\vec{X} \leftarrow \vec{x}'$ constitutes a candidate actual cause-witness pair.
    Note that the set of interventions encoded in any descendant of the pruned \textit{Agent} node is either identical to $\vec{X} \leftarrow \vec{x}'$ apart from its last counterfactual action, or its variable set is a superset of $\vec{X}$, and hence it is non-minimal according to Definition \ref{def.ac}. 
    \item It is an \textit{Agent} node whose encoded set of interventions is non-minimal w.r.t. to the current set of found candidate actual cause-witness pairs.
    \item It is a fully-expanded node with all of its children already pruned.
\end{itemize}

\subsection{Responsibility Attribution Using Monte Carlo Tree Search (RA-MCTS)}\label{sec.mcts}

The search algorithm we propose is based on the well-known Monte Carlo Tree Search (MCTS) method \cite{browne2012survey}. We refer to our algorithm as RA-MCTS because it is specific to the task of responsibility attribution.
The main differences between RA-MCTS and standard MCTS \cite{coulom2006efficient, chaslot2008monte} are in their \textit{simulation phases}, \textit{evaluation functions}, \textit{child selection policies} and \textit{backpropagation phases}.

\textit{\textbf{Simulation Phase.}} 
At each iteration, the entire simulation path is added to the search tree. Although in applications of MCTS the tree is usually expanded by one node per iteration, this would not be optimal in our setting. Namely, under a fixed causal setting, the state transitions, observations generation and other such functions are deterministic.\footnote{Similar MCTS modifications have been used in other deterministic tasks, such as guiding symbolic execution in generating useful visual programming tasks \cite{ahmed2020synthesizing}.} Hence, computing their values more than once is a waste of computational resources.

\textit{\textbf{Evaluation Function.}}
Whenever a \textit{Leaf} node $L$ is visited during an iteration of (RA-)MCTS, a score is assigned to it and then backpropagated to all of its ancestors. Properly defining the function that determines that score, i.e., the \textit{evaluation function}, is considered to be a critical ingredient of successfully applying MCTS methods.
Considering the idiosyncrasy of our task, we design an evaluation function that returns a multi-dimensional score, as opposed to a single numerical value which is typically the case.
More precisely, this evaluation function takes as input the set of interventions $\vec{X} \leftarrow \vec{x}'$ encoded in $L$, and outputs a score vector $\vec{r}$, of size $n+1$, which is defined as follows.
For each agent $i \in \{1, ..., n\}$, $r_i = dr_i((\mathcal{C}, \vec{u}), \phi^\tau, \vec{X} \leftarrow \vec{x}', \mathcal{D})$, where $\mathcal{D}$ denotes Definition \ref{def.ac}.
The $n+1$th value of $\vec{r}$ is equal to the output of an environment specific function $q_{\mbox{env}}((\mathcal{C}, \vec{u}), \vec{X} \leftarrow \vec{x}')$ which provides some additional information about the final outcome that corresponds to the causal setting $(\mathcal{C}^{\vec{X} \leftarrow \vec{x}'}, \vec{u})$.\footnote{If such a function is not available then this part can be omitted.}
For example, in a card game scenario we typically want to attribute responsibility to the members of the team that lost (outcome). Additional information that our search algorithm could benefit from in this scenario is how closer to or further from winning would the losing team get, had we intervened on some actions taken by its members.
The purpose of $\vec{r}$'s first $n$ values is to guide the search towards optimizing its main objective, i.e., approximating the agents' degrees of responsibility. 
The role of the last value of $\vec{r}$ is complementary, as it helps to identify areas which are promising for discovering new actual causes.

\textit{\textbf{Child Selection Policy.}}
Similar to standard MCTS, in RA-MCTS each node $v$ keeps track of two statistics, the number $N(v)$ of times it has been visited and the vector $\vec{Q}(v)$, where $Q_j(v)$, with $j \in \{1, .., n+1\}$, is equal to the total score $r_j$ of all simulations that passed through that node.
In order to transform $\vec{Q}(v)$ into a scalar score value, the child selection policy of RA-MCTS follows a \textit{linear scalarization} approach inspired by the Multi-Objective Multi-Armed Bandits (MO-MAB) literature \cite{drugan2013designing, tekin2018multi}.
At iteration $k$, a pre-defined weight $b_{k, j}$ is assigned to each value of $\vec{Q}(v)$, where $\sum_{j \in \{1, ..., n+1\}} b_{k, j} = 1$. The linear scalarized value of $\vec{Q}(v)$ at iteration $k$ is then defined as
\begin{align}\label{eq.lsf}
    f_{LS}(\vec{Q}(v)) = \sum_{j \in \{1, ..., n+1\}} b_{k, j} \cdot Q_j(v).
\end{align}
After computing $f_{LS}(\vec{Q}(v))$ for each child node $v$, our policy employs the UCB$1$ formula \cite{auer2002finite, kocsis2006bandit} to select the next node in the path

\begin{align}\label{eq.ucb}
    v^{next} := \argmax_{v \in \mbox{children}} \frac{f_{LS}(\vec{Q}(v))}{N(v)} + C \cdot \sqrt{\frac{\ln{N}}{N(v)}},
\end{align}
where $N$ is the parent node's number of visitations and $C$ is the \textit{exploration} parameter of RA-MCTS. Ties are broken randomly.

In our experiments, at every iteration $k$, we set $b_{k, n+1} := B$, where $B \in [0, 1)$ is a constant.
Additionally, for $i = (k \;\mathrm{mod}\; n)$ we set $b_{k, i} := 1 - B$, while every other weight $b_{k,j}$, with $j \notin \{i, n+1\}$, is set to $0$.
As a result, the only simulated responsibility degrees that guide the search at iteration $k$ are those of agent $i$.

\textit{\textbf{Backpropagation Phase.}} 
Note that as the set of found candidate actual cause-witness pairs grows during search, the intervention set encoded in a previously expanded \textit{Agent} node $v$ might be evaluated as non-minimal when $v$ gets visited again.
Whenever this happens, in addition to pruning $v$, we also backpropagate values $(-\vec{Q}(v), -N(v))$ to its ancestors.
This way, we completely erase the footprints of the pruned node from the rest of the tree.
Therefore, by taking this measure our search method is no longer guided by scores of simulations that passed through $v$.

\subsection{Estimating Responsibility Assignments under Context Uncertainty}\label{sec.alg_unc}

Our analysis  so far in this section, assumes context $\vec{u}$ to be known. We now lift this assumption and propose our solution to the inference problem described in Section \ref{sec.prob_chall}. We extend RA-MCTS in the following way.
First, we draw $M$ Monte Carlo samples from the posterior $Pr(\vec{u}|\tau)$, utilizing the procedure described in Section 3.4 of \cite{oberst2019counterfactual}.
Next, we compute for each agent $i$ its average degree of responsibility over all samples
\begin{align}\label{eq.mean_resp}
    \overline{d}_i := \frac{1}{M} \cdot \sum_{m \in \{1, ..., M\}} d^m_i,
\end{align}
where $d^m_i$ is $i$'s degree of responsibility in $(\mathcal{C}, \vec{u}_m)$ according to RA-MCTS, and $\vec{u}_m$ is the $m$th sample.


\section{Experiments}\label{sec.experiments}

In this section, we experimentally test the efficacy of RA-MCTS, for known and unknown context, using a simulation-based testbed, which contains three card games. 
In our experiments, we restrict the maximum size of actual cause-witness pairs to $4$, for reasons explained in \cite{10.1145/3514094.3534133}. 
We also fix RA-MCTS parameters to $C=2$ and $B=0.5$.
Additional results can be found in \iftoggle{arxiv}{Appendices \ref{app.ablation} and \ref{app.unc}}{the Supplemental Material}.

\captionsetup[figure]{belowskip=-10pt}
\begin{figure*}
\captionsetup[subfigure]{aboveskip=0pt,belowskip=1pt}
\centering
    \begin{subfigure}[c]{0.5\textwidth}
        \includegraphics[width=\textwidth]{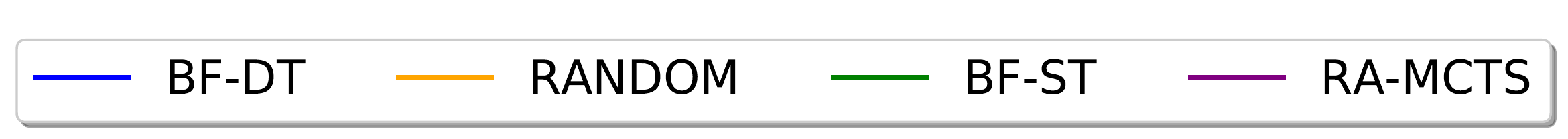}
    \end{subfigure}\\
    \begin{subfigure}[c]{0.24\textwidth}
        \includegraphics[width=\textwidth]{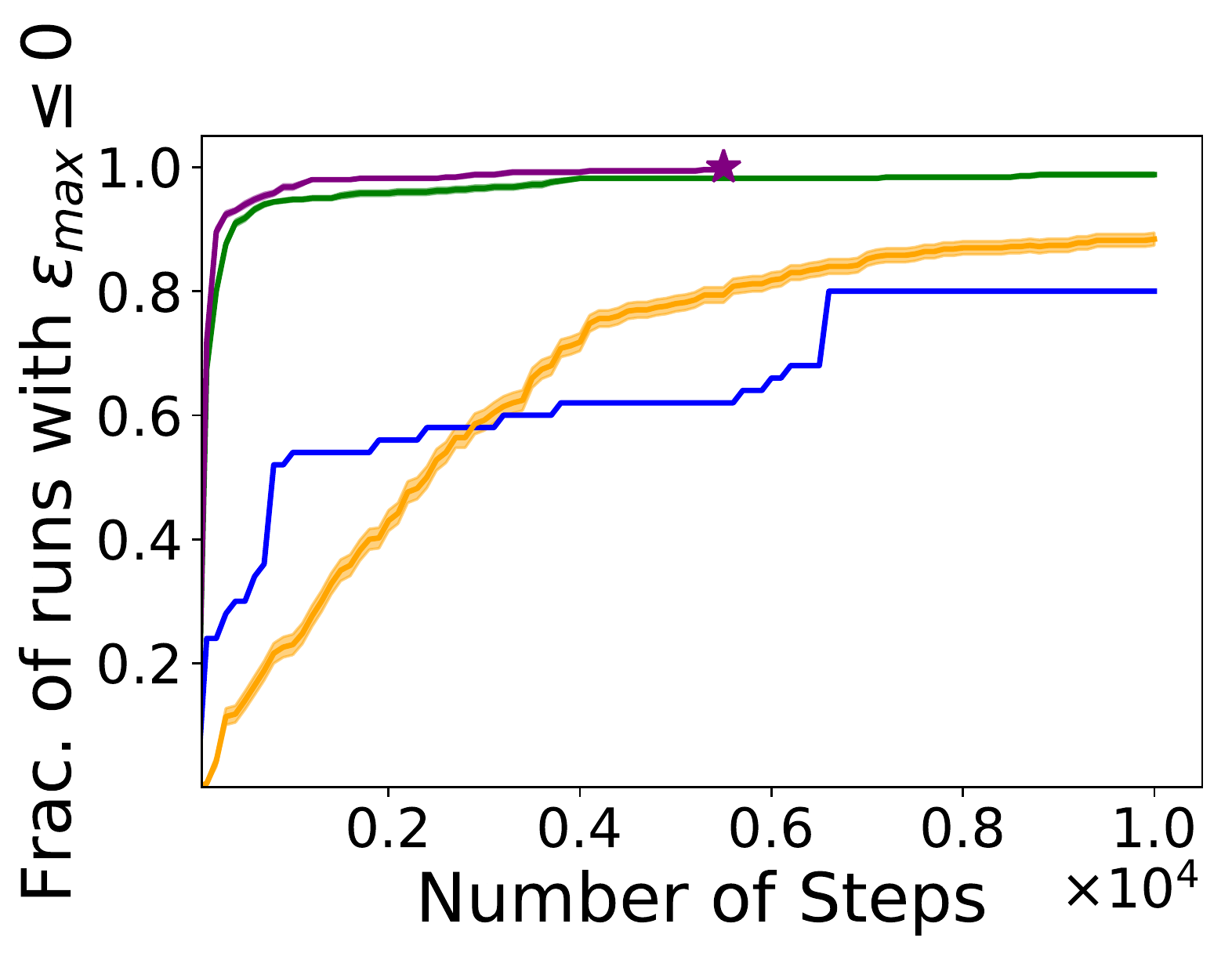}
        \captionsetup{type=figure}
        \caption{\textit{TeamGoofspiel}$\boldsymbol{(7)}$}
        \label{fig : no_unc_team_goofspiel_7}
    \end{subfigure}\hfill%
    \begin{subfigure}[c]{0.24\textwidth}
        \includegraphics[width=\textwidth]{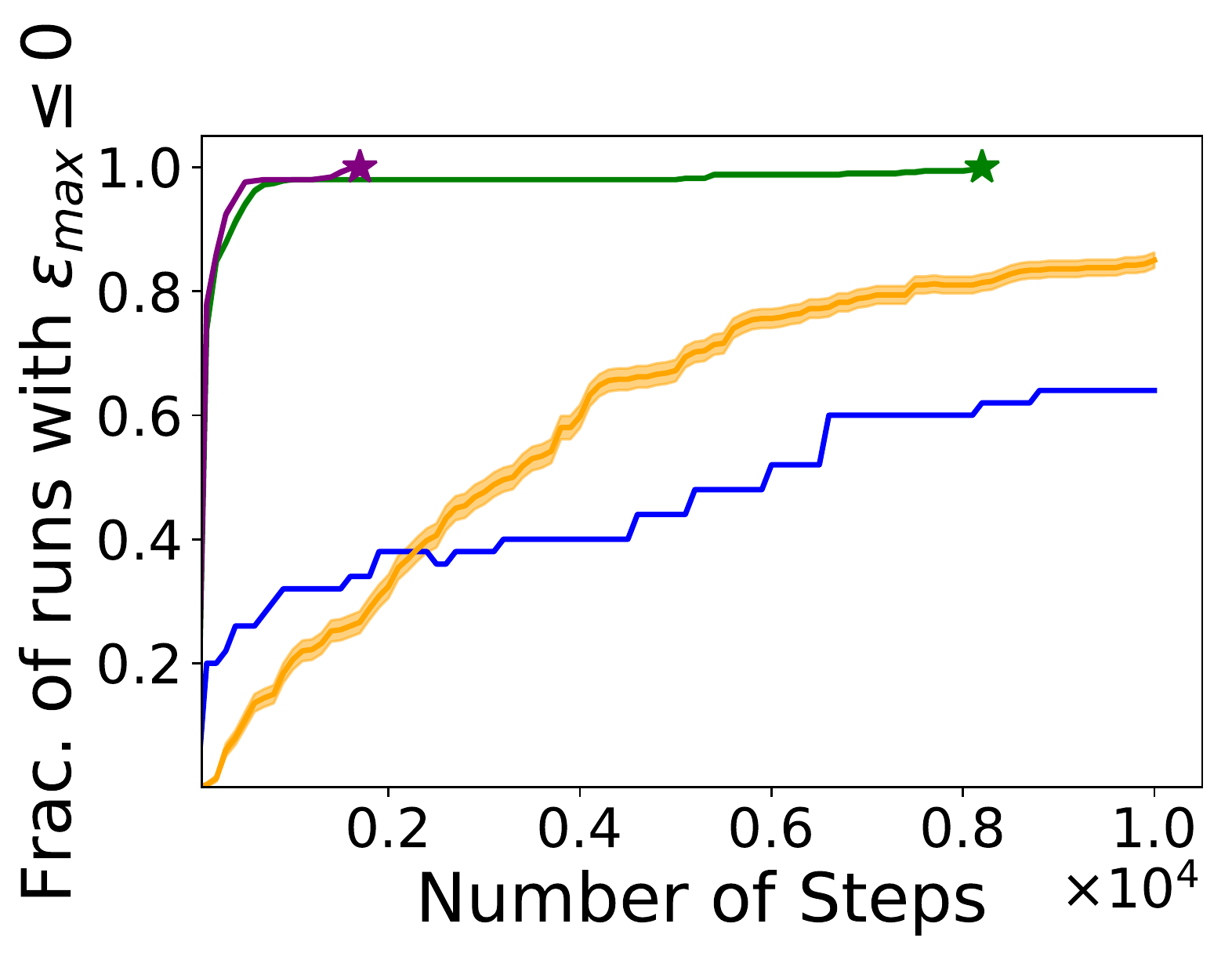}
        \captionsetup{type=figure}
        \caption{\textit{TeamGoofspiel}$\boldsymbol{(8)}$}
        \label{fig : no_unc_team_goofspiel_8}
    \end{subfigure}\hfill%
    \begin{subfigure}[c]{0.24\textwidth}
        \includegraphics[width=\textwidth]{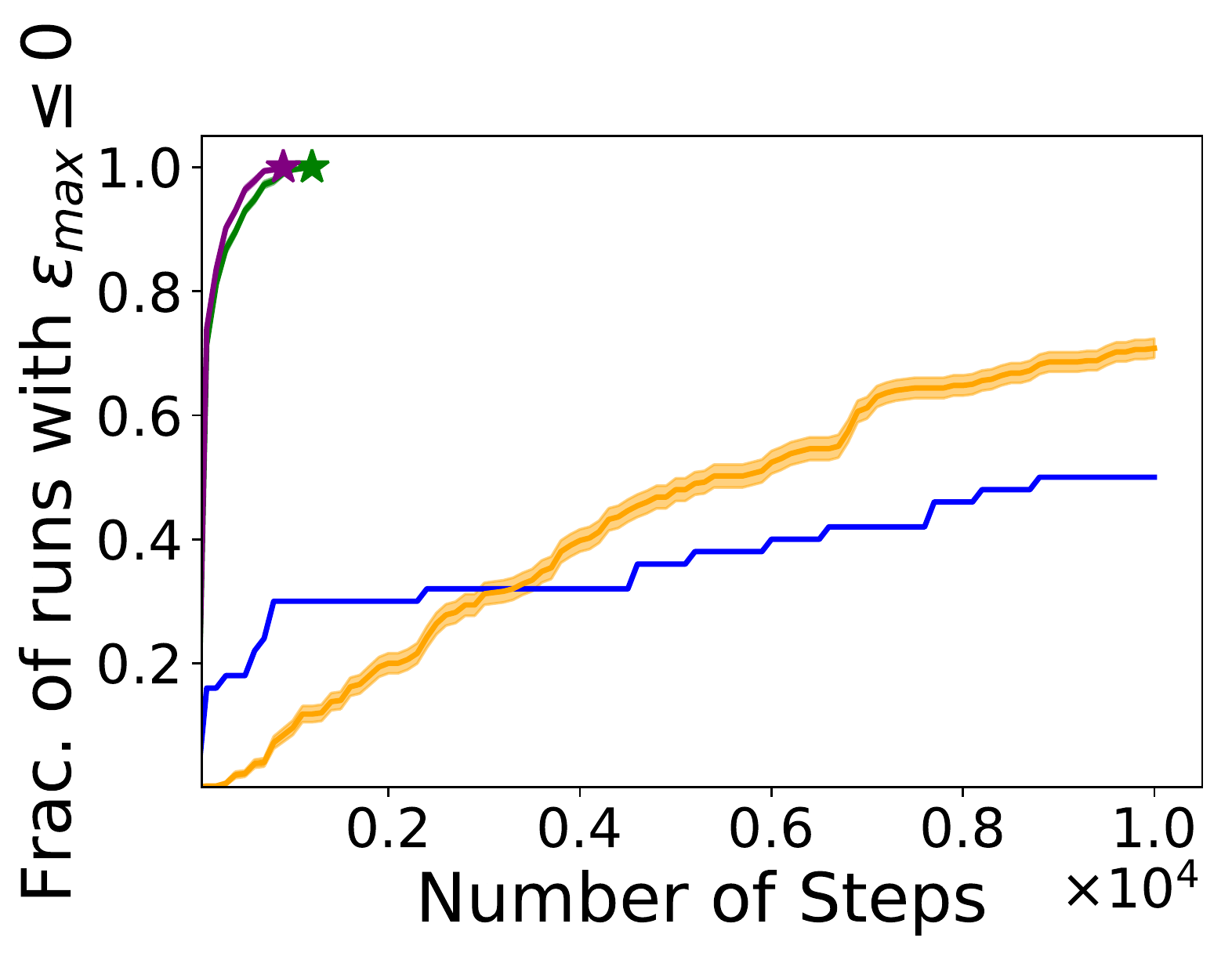}
        \captionsetup{type=figure}
        \caption{\textit{TeamGoofspiel}$\boldsymbol{(9)}$}
        \label{fig : no_unc_team_goofspiel_9}
    \end{subfigure}
    \begin{subfigure}[c]{0.24\textwidth}
        \includegraphics[width=\textwidth]{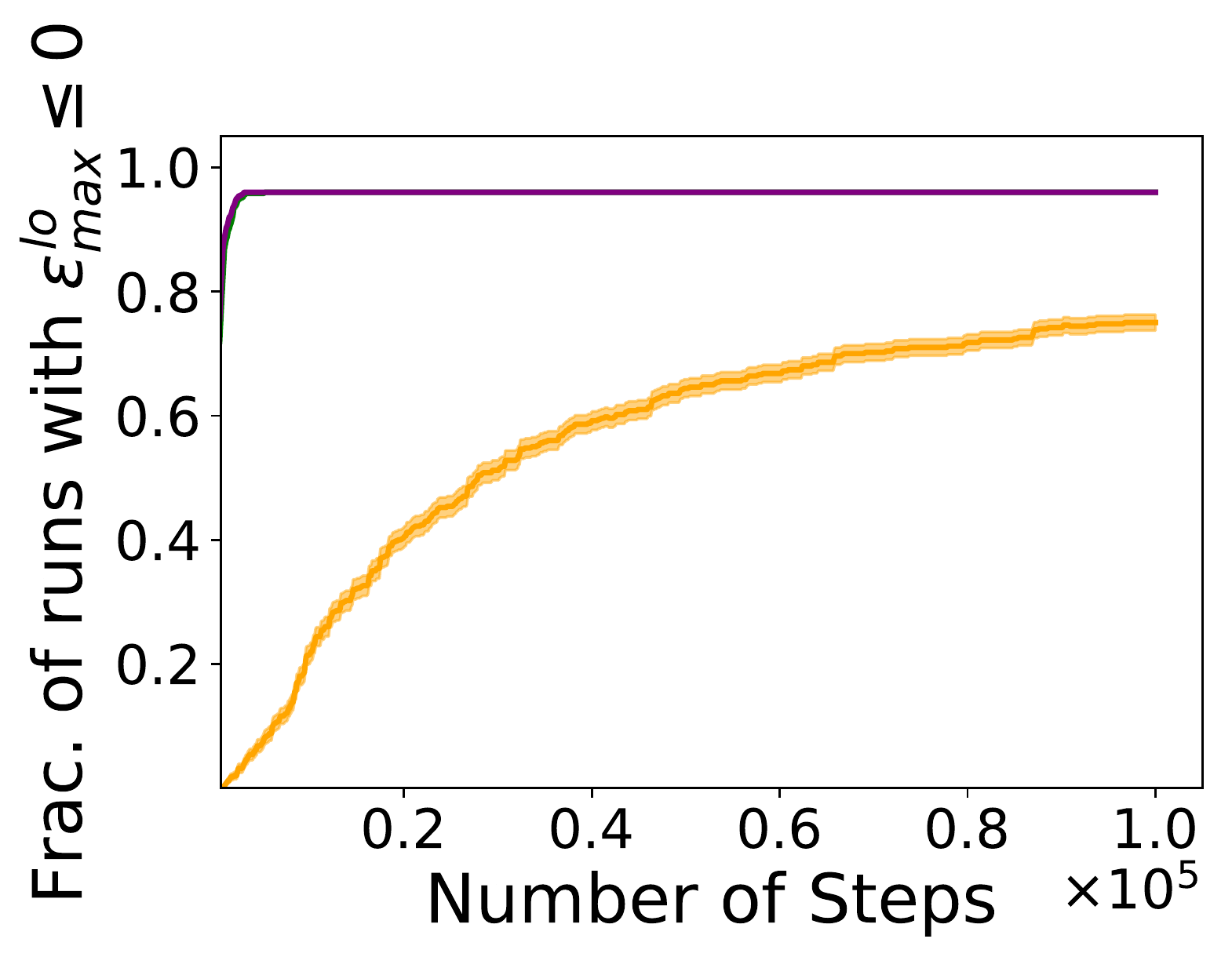}
        \captionsetup{type=figure}
        \caption{\textit{TeamGoofspiel}$\boldsymbol{(13)}$}
        \label{fig : no_unc_team_goofspiel_13}
    \end{subfigure}\\
    \begin{subfigure}[c]{0.24\textwidth}
        \includegraphics[width=\textwidth]{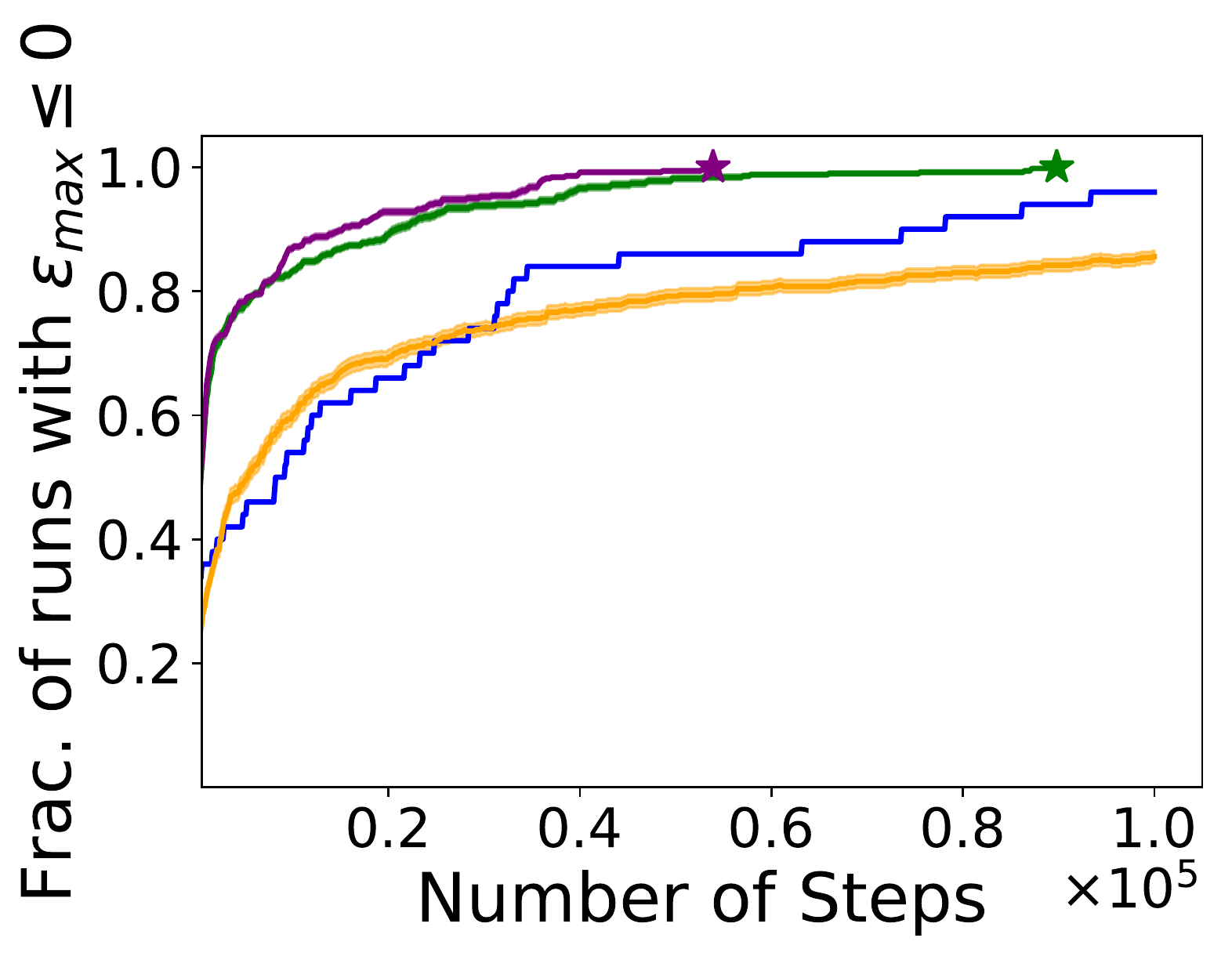}
        \captionsetup{type=figure}
        \caption{\textit{Euchre}$\boldsymbol{(8)}$}
        \label{fig : no_unc_euchre_8}
    \end{subfigure}\hfill%
    \begin{subfigure}[c]{0.24\textwidth}
        \includegraphics[width=\textwidth]{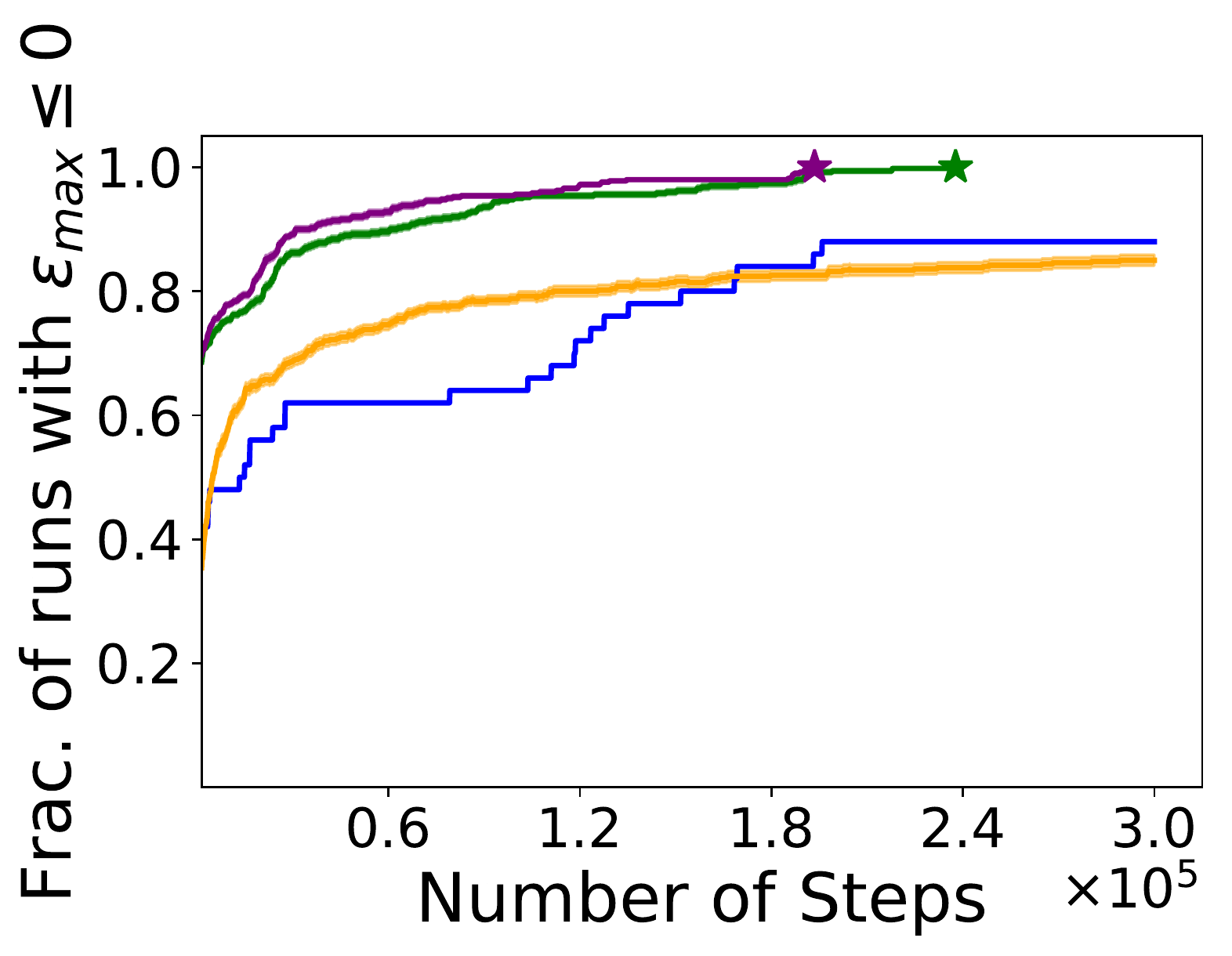}
        \captionsetup{type=figure}
        \caption{\textit{Euchre}$\boldsymbol{(9)}$}
        \label{fig : no_unc_euchre_9}
    \end{subfigure}\hfill%
    \begin{subfigure}[c]{0.24\textwidth}
        \includegraphics[width=\textwidth]{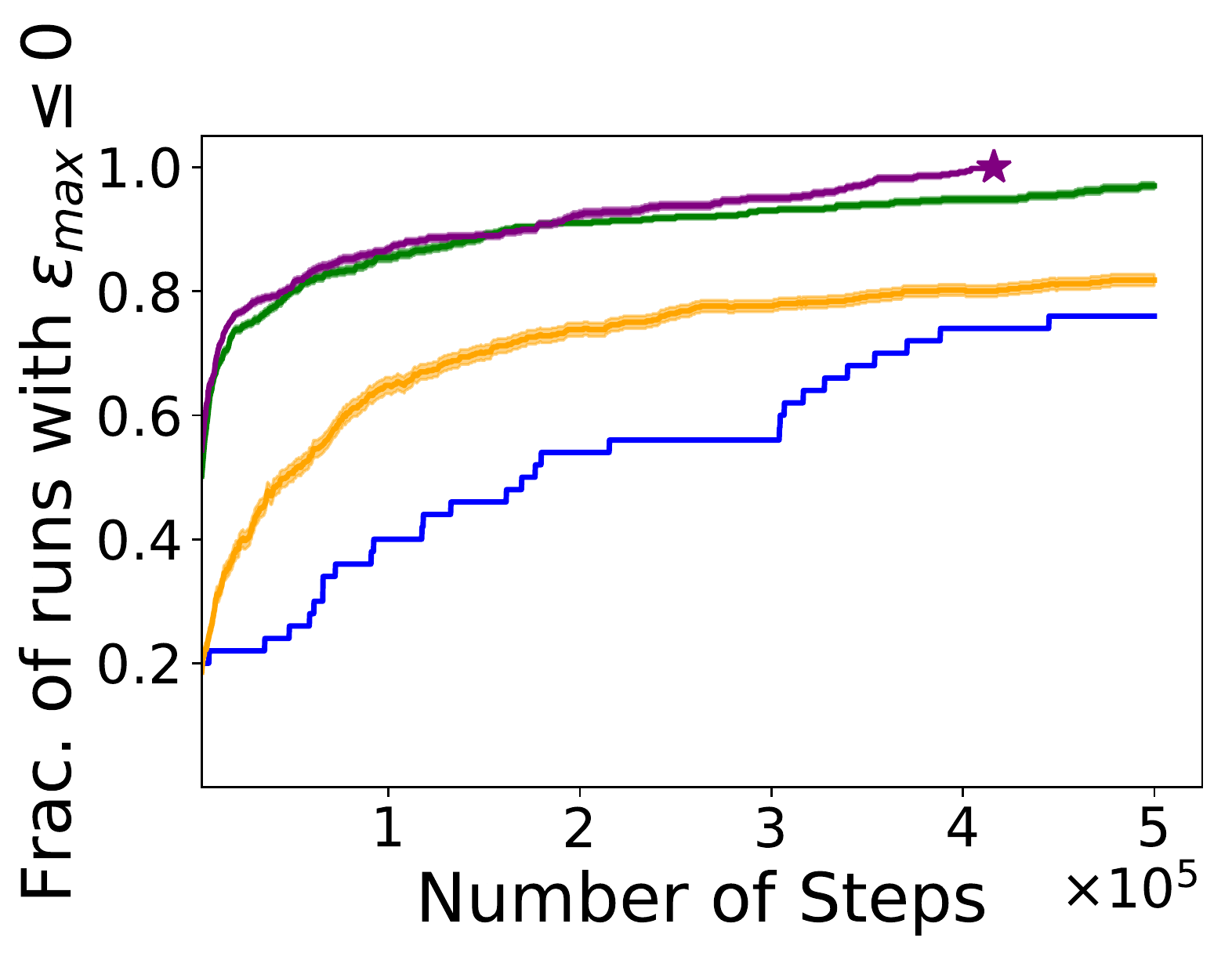}
        \captionsetup{type=figure}
        \caption{\textit{Euchre}$\boldsymbol{(10)}$}
        \label{fig : no_unc_euchre_10}
    \end{subfigure}\hfill%
    \begin{subfigure}[c]{0.24\textwidth}
        \includegraphics[width=\textwidth]{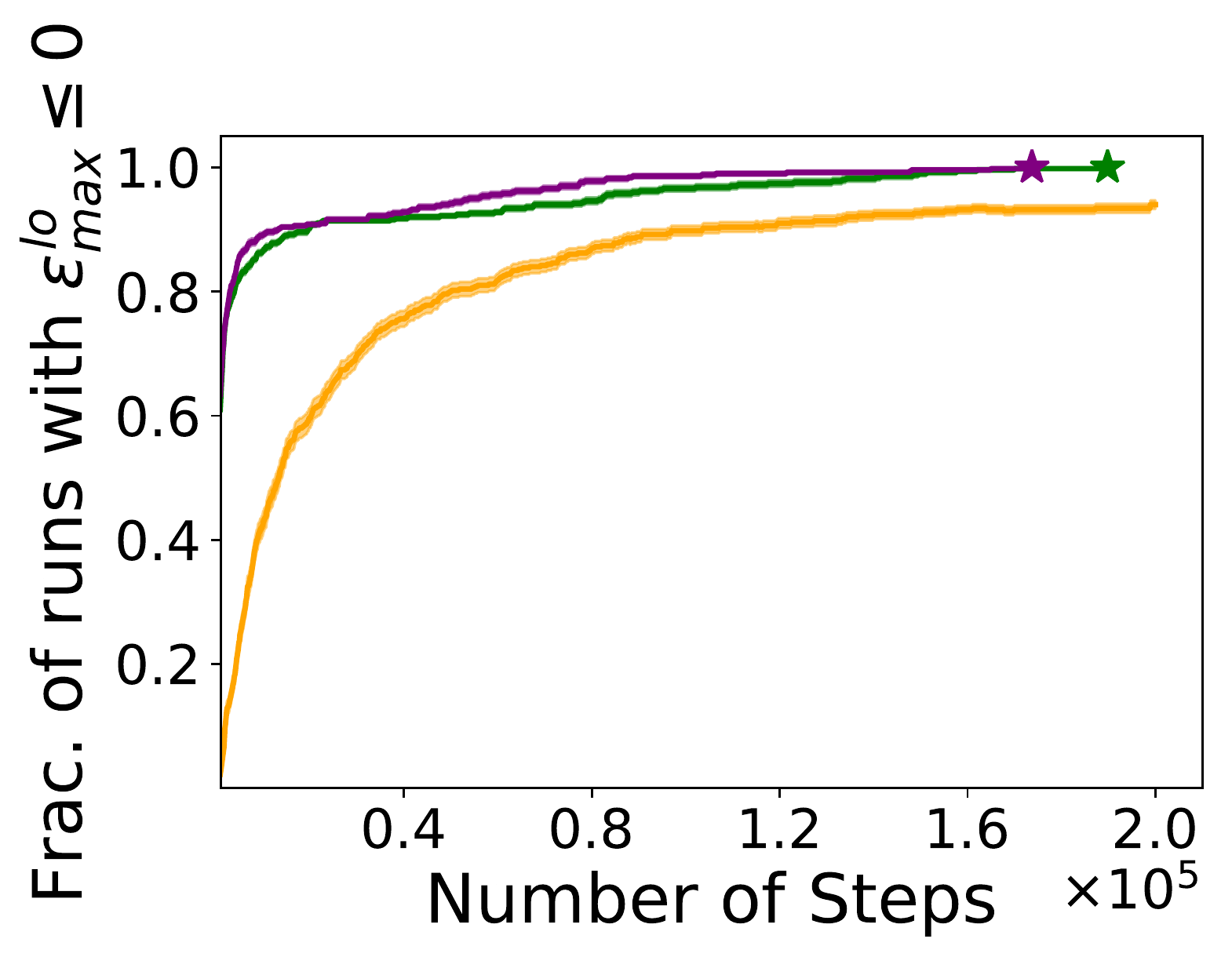}
        \captionsetup{type=figure}
        \caption{\textit{Euchre}$\boldsymbol{(12)}$}
        \label{fig : no_unc_euchre_12}
    \end{subfigure}\\
    \begin{subfigure}[c]{0.24\textwidth}
        \includegraphics[width=\textwidth]{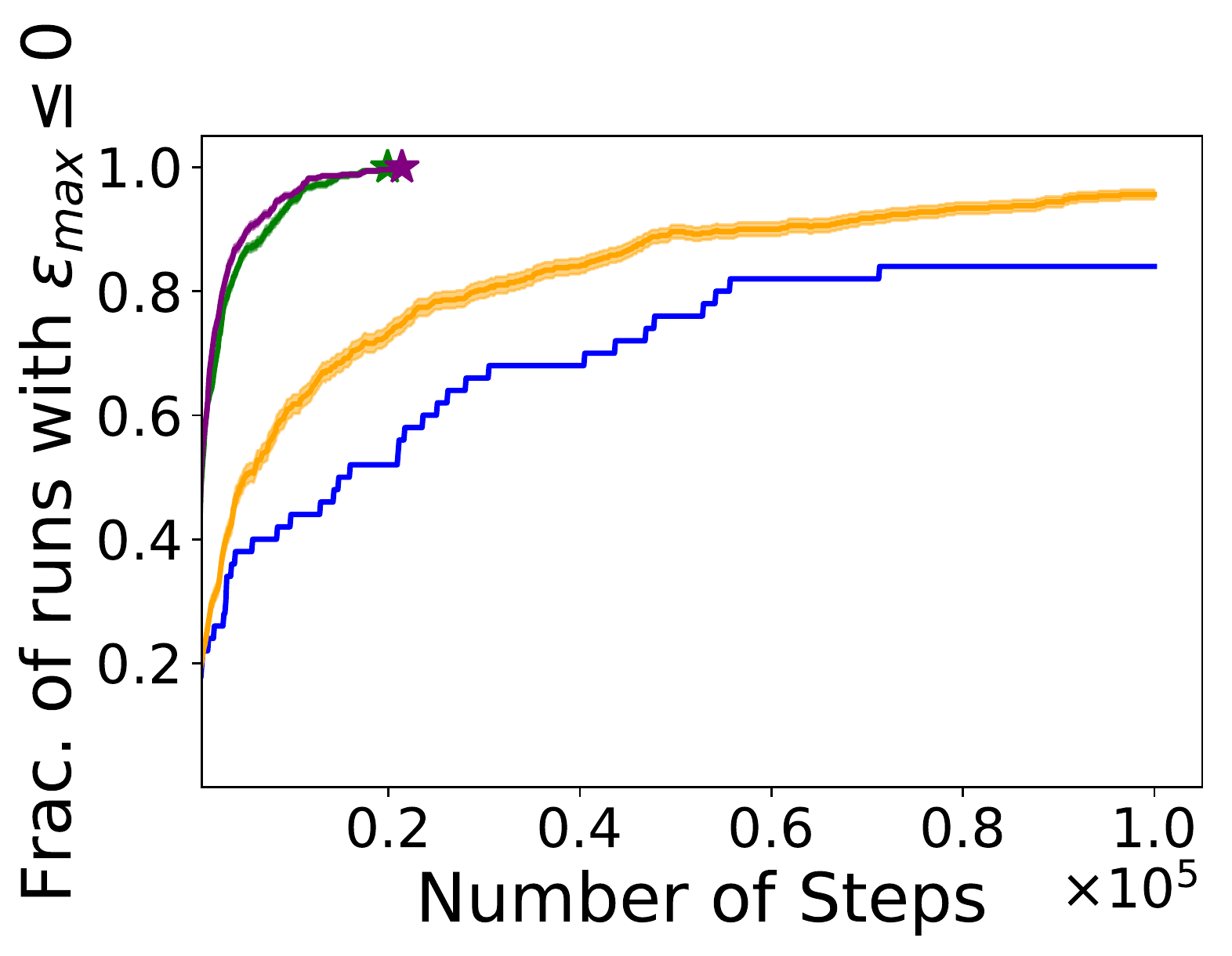}
        \captionsetup{type=figure}
        \caption{\textit{Spades}$\boldsymbol{(8)}$}
        \label{fig : no_unc_spades_8}
    \end{subfigure}\hfill%
    \begin{subfigure}[c]{0.24\textwidth}
        \includegraphics[width=\textwidth]{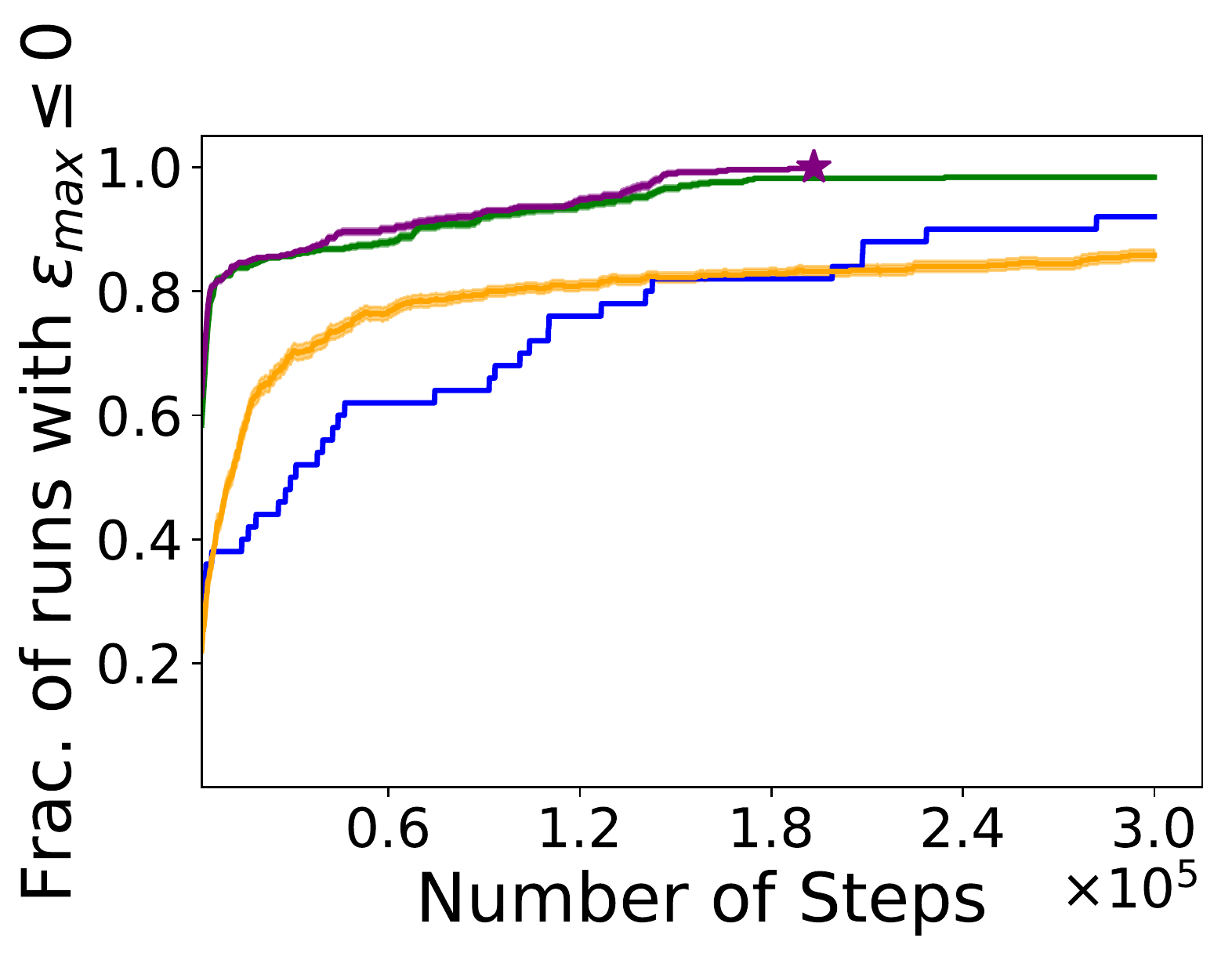}
        \captionsetup{type=figure}
        \caption{\textit{Spades}$\boldsymbol{(9)}$}
        \label{fig : no_unc_spades_9}
    \end{subfigure}\hfill%
    \begin{subfigure}[c]{0.24\textwidth}
        \includegraphics[width=\textwidth]{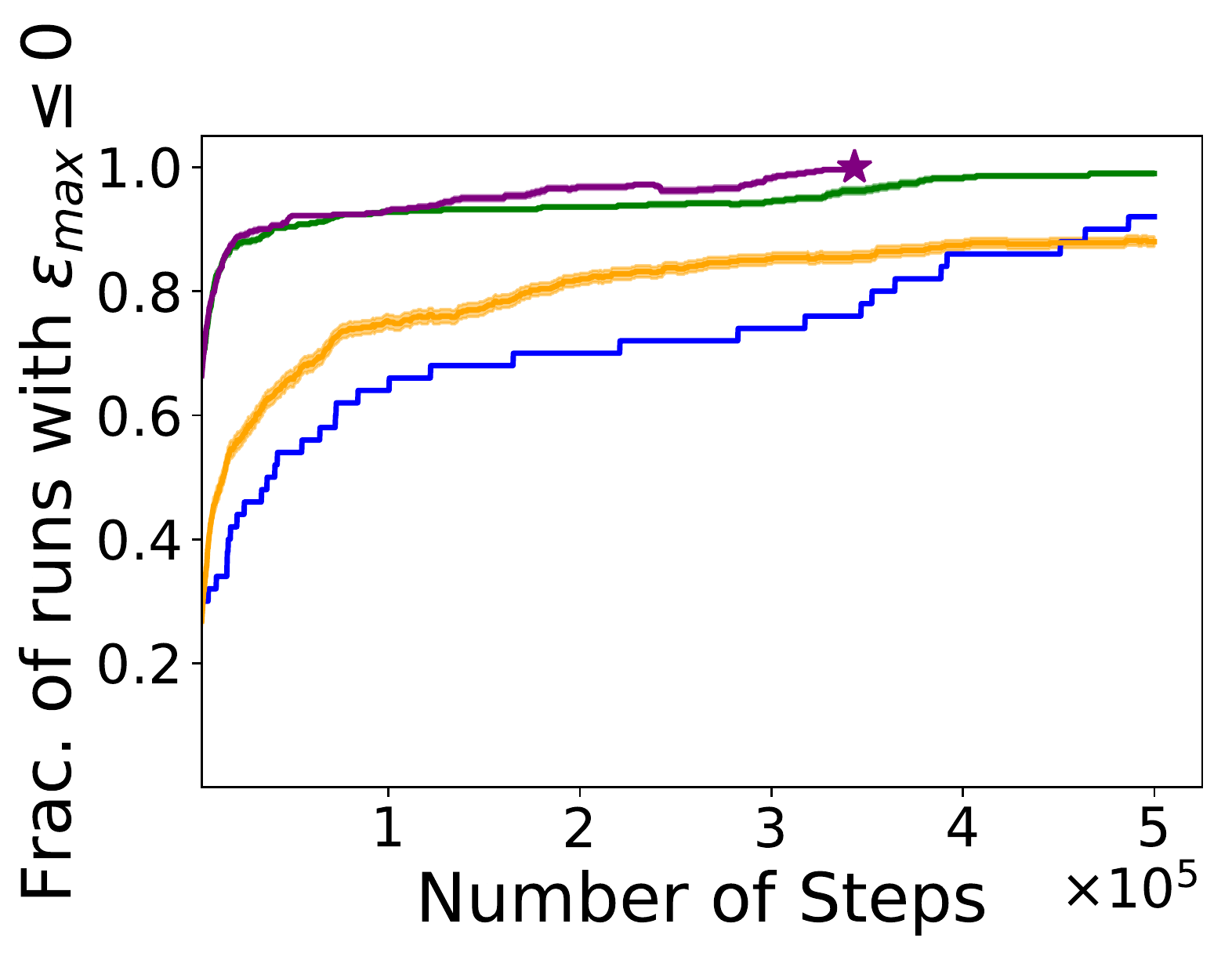}
        \captionsetup{type=figure}
        \caption{\textit{Spades}$\boldsymbol{(10)}$}
        \label{fig : no_unc_spades_10}
    \end{subfigure}\hfill%
    \begin{subfigure}[c]{0.24\textwidth}
        \includegraphics[width=\textwidth]{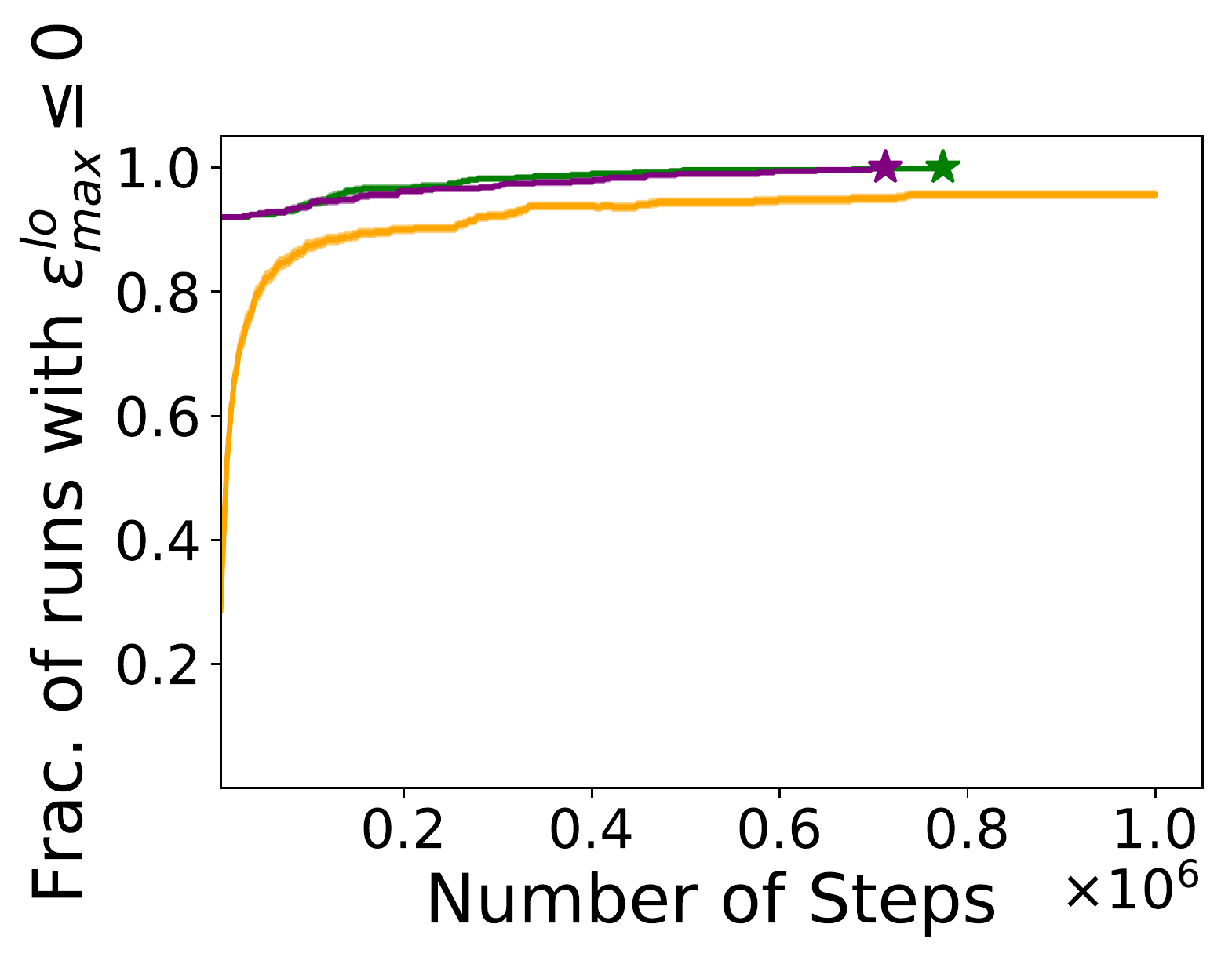}
        \captionsetup{type=figure}
        \caption{\textit{Spades}$\boldsymbol{(13)}$}
        \label{fig : no_unc_spades_13}
    \end{subfigure}
\caption{Performance profiles on \textit{TeamGoofspiel}, \textit{Euchre} and \textit{Spades} with known context. Shaded regions show standard deviation. We add a marker if at the current number of steps the fraction of runs with $\epsilon_{max} \leq 0$ or $\epsilon^{lo}_{max} \leq 0$ is $1$.} 
\label{fig: plots}
\end{figure*}

\subsection{Environments and Policies}\label{sec.envs_policies}

We consider three card games played by two teams of two players. 
The members of one team are referred to as opponents, and they are treated as part of the environment. The members of the other team are treated as agents, and they are denoted by \agentzero~and \agentone.
All players have the same (initial) information probability function, but different decision making policies.

\textbf{\textit{TeamGoofspiel}$\boldsymbol{(H)}$.} The first game is a team variation of the card game Goofspiel, introduced in \cite{10.1145/3514094.3534133}. In this game, the initial hand of each player consists of $H$ cards. Typically, $H=13$.
At each round, all players simultaneously discard one of their cards, after observing the round's prize. The team which played the cards with highest total value collects the prize. After $H$ rounds, the team that accumulated the biggest prize wins the game. 
Agent \agentzero~tries to always play the card whose value matches the round's prize. If that card is not in \agentzero's hand, then it chooses a card based on which team is currently leading the game. Agent \agentone~chooses its card based on a comparison between the average value of its hand and the current round's prize.
Opponents follow the same stochastic policy which assigns a distribution on their hand based on the round's prize and the current leading team.
For more details on the rules of the game and the players' policies see \cite{10.1145/3514094.3534133}.

\textbf{\textit{Euchre}$\boldsymbol{(H)}$.} Second, we consider a turn-based trick-taking game. Each player is initially dealt $H$ cards from a standard deck, with $H$ typically being $5$.
Next follows the \textit{calling} phase, where the \textit{trump suit} and the player who starts first are chosen. For simplicity, we omit this phase and make the aforementioned choices randomly.
At each round, the first player discards one card. This card's suit becomes the \textit{leading suit} of the current round. The rest of the players (in clockwise order) have to follow the \textit{leading suit} if possible, otherwise they are allowed to play any card from their hand. The winner of the round is determined by a game-specific \textit{card ranking} which takes into account the trump and the lead suits. The player who won the previous round starts next.
After $H$ rounds, the team with the most wins takes the game.
The policies of agents \agentzero~and \agentone~are based on the \textit{HIGH!} policy \cite{seelbinder2012cooperative}. The main idea of \textit{HIGH!} is that ``\textit{if your teammate leads the round then let them win}''. We implement the policy of \agentzero~to be slightly more aggressive than that of \agentone.
Opponents' policies follow the \textit{HIGH!} principle only when they play last in a round, otherwise they follow a stochastic greedy policy which assigns higher probabilities to cards that have potential to win the round. For more information see \iftoggle{arxiv}{Appendix \ref{app.euchre}}{the Supplemental Material}.

\textbf{\textit{Spades}$\boldsymbol{(H)}$.}
Our third card game is yet another trick-taking game which is similar to \textit{Euchre}$(H)$, but with some key differences.
For example, there is no calling phase and the trump suit is always spades. Before they start playing, the players must \textit{bid} on the number of tricks they believe that they will have won after $H$ rounds, where typically $H=13$.
\textit{Spades}$(H)$ has a different card ranking than \textit{Euchre}$(H)$, and also some additional rules on which cards are allowed to be discarded by the players at each time.
At the end of the game, the score of each team is calculated based on the number of tricks it won and its initial bids. In case a team bid more than its won tricks, it receives a penalty based on a \textit{sandbagging rule}.
The players' policies in \textit{Spades}$(H)$ are very similar to the ones in \textit{Euchre}$(H)$. For more information see \iftoggle{arxiv}{Appendix \ref{app.spades}}{the Supplemental Material}.

Note that all three games are standard benchmarks for AI research \cite{hennes2020neural, kaur2005fuzzy, cowling2014player, baier2018emulating}, and have also received extensive mathematical analysis \cite{ross1971goofspiel, rhoads2012computer, grimes2013observations, wright2001mathematics, cohensius2019bidding}. Moreover, \textit{Goofspiel} and \textit{Euchre} are parts of a well known framework for RL in games \cite{lanctot2019openspiel}.

\subsection{Experimental Setup}\label{sec.setup}
\captionsetup[figure]{belowskip=-10pt}
\begin{figure*}
\captionsetup[subfigure]{aboveskip=0pt,belowskip=1pt}
\centering
    \begin{subfigure}[c]{1\textwidth}
        \includegraphics[width=\textwidth]{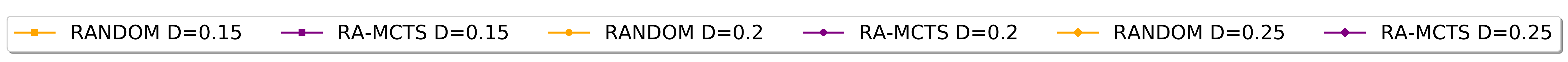}
    \end{subfigure}\\
    \begin{subfigure}[c]{0.24\textwidth}
        \includegraphics[width=\textwidth]{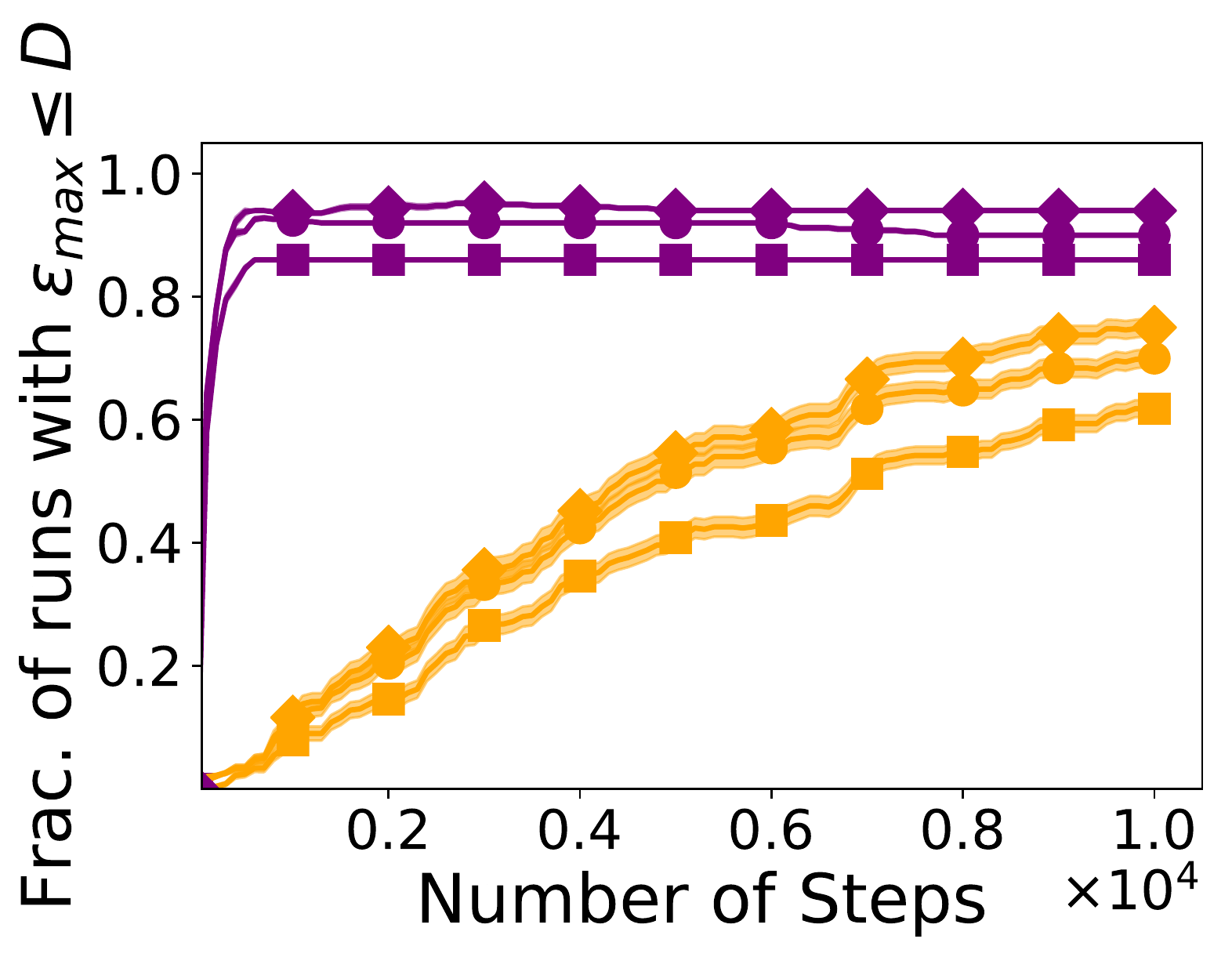}
        \captionsetup{type=figure}
        \caption{\textit{TeamGoofspiel}$\boldsymbol{(9)}$ Unc.}
        \label{fig : unc_team_goofspiel_9}
    \end{subfigure}\hfill%
    \begin{subfigure}[c]{0.24\textwidth}
        \includegraphics[width=\textwidth]{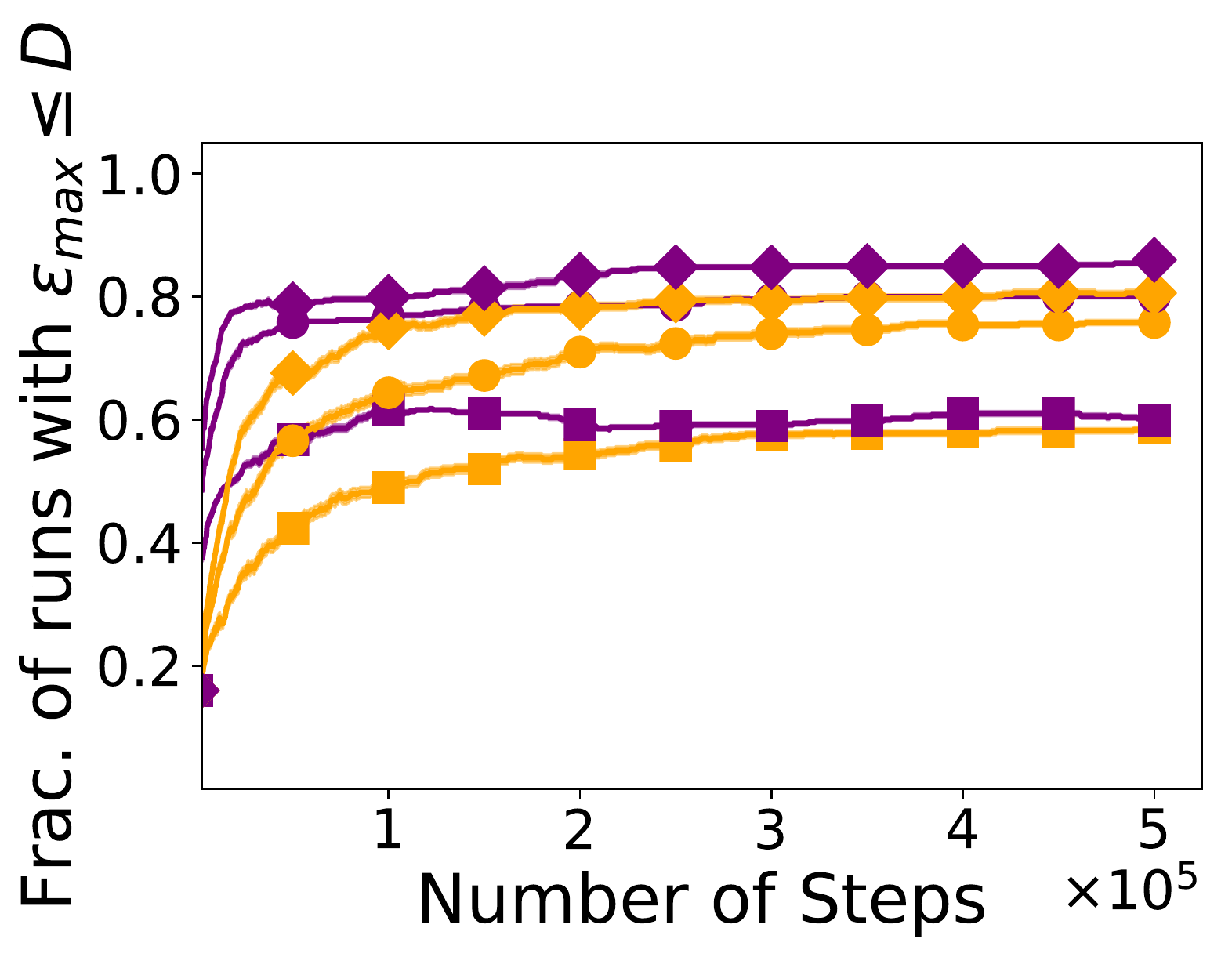}
        \captionsetup{type=figure}
        \caption{\textit{Euchre}$\boldsymbol{(10)}$ Unc.}
        \label{fig : unc_euchre_10}
    \end{subfigure}\hfill%
    \begin{subfigure}[c]{0.24\textwidth}
        \includegraphics[width=\textwidth]{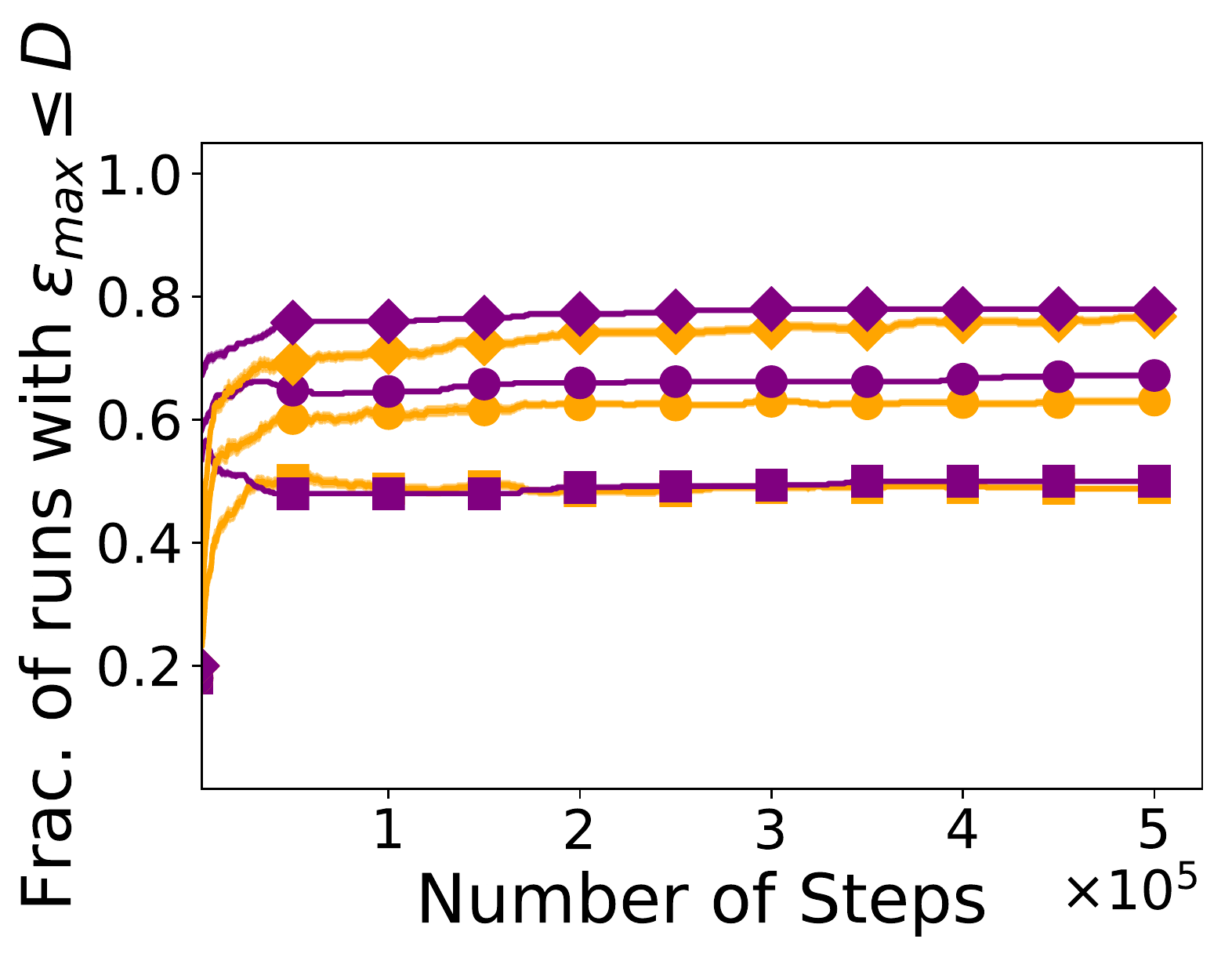}
        \captionsetup{type=figure}
        \caption{\textit{Spades}$\boldsymbol{(10)}$ Unc.}
        \label{fig : unc_spades_10}
    \end{subfigure}
\caption{Performance profiles on \textit{TeamGoofspiel}$\boldsymbol{(9)}$, \textit{Euchre}$\boldsymbol{(10)}$ and \textit{Spades}$\boldsymbol{(10)}$ with unknown context. Shaded regions show standard deviation and number of steps are per sample.
The full set of plots can be found in \iftoggle{arxiv}{Appendix \ref{app.unc}}{the Supplemental Material}.}
\label{fig: unc_plots}
\end{figure*}

We evaluate the efficacy of several search algorithms on estimating a responsibility assignment under a computational budget. Computational budget in our experiments is defined as the total number of environment steps that an algorithm is allowed to take.

\textbf{Baselines.} Apart from RA-MCTS we also implement RANDOM, which repeatedly samples a set of interventions and checks whether it constitutes a candidate actual cause-witness pair or not. When computational budget is reached, RANDOM determines the agents' degrees of responsibility based on the found solutions.
Other baselines are BF-DT and BF-ST, which perform a brute force search over all possible sets of interventions. BF-DT is the algorithm of choice in \cite{10.1145/3514094.3534133}, and it utilizes the standard decision (game) tree.
On the other hand, BF-ST utilizes the search tree from Section \ref{sec.tree}.

\textbf{Performance profiles.}
We generate multiple configurations of our environments by changing parameter $H$. For each of these configurations, our methods are evaluated on $50$ different trajectories, in which the agents fail to win the opponents. For each such trajectory, we perform $10$ independent runs of each method.
\footnote{We change the initial seed of the method.}
Following \cite{agarwal2021deep}, we report performance profiles based on \textit{run-score distributions}, and show for each method the fraction of runs in which it performs better than a certain threshold $D$.
We measure the performance of a method in terms of its accuracy w.r.t. some target responsibility assignment. 
More specifically, when it is computationally feasible to find the exact responsibility assignment, we report the maximum absolute difference $\epsilon_{max} \in [0,1]$. For instance, if for some trajectory the agents' degrees of responsibility according to a method are $0.25$ and $0.75$, but the exact degrees are $0.33$ and $1$, then $\epsilon_{max} = 0.25$.
If instead of the exact values, we can only compute lower bounds of the agents' responsibilities, we report the maximum absolute lower difference $\epsilon^{lo}_{max} \in [0,1]$. Going back to our previous example, if the known lower bounds are $0.33$ and $0.5$, then $\epsilon^{lo}_{max} = 0.08$.

For \textit{Euchre}$(H)$ and \textit{Spades}$(H)$, we are able compute the exact responsibility assignments for values of $H$ up to $10$.\footnote{They are found using BF-DT, as in \cite{10.1145/3514094.3534133}.} For \textit{TeamGoofspiel}$(H)$, the upper limit is $9$. In order to evaluate our methods on environments with larger $H$, we follow a procedure, described in \iftoggle{arxiv}{Appendix \ref{app.lower}}{the Supplemental Material}, and generate trajectories for which we can retrieve non-trivial lower bounds of the agents' degrees of responsibility.

\subsection{Results}\label{sec.results}

\subsubsection{Results with Known Context}\label{sec.no_unc_results}

Plots \ref{fig : no_unc_team_goofspiel_7}-\ref{fig : no_unc_spades_13} display performance profiles for the implemented methods and for different computational budgets. Note that threshold $D$ in these experiments is set to $0$. This means that a method performs better than $D$ iff it manages to find the exact responsibility assignment.

We observe that RA-MCTS almost always converges to the optimal solution, i.e., achieves $\epsilon_{max} = 0$ or $\epsilon^{lo}_{max} = 0$, under a reasonable computational budget. The only exception is \textit{TeamGoofspiel}$(13)$, for which it converges for $96\%$ of the runs.
To better understand how budget efficient RA-MCTS is, consider configurations \textit{Euchre}$(10)$ and \textit{Spades}$(10)$.
By looking at Plots \ref{fig : no_unc_euchre_10} and \ref{fig : no_unc_spades_10}, we observe that RA-MCTS needs at most $4.2 \cdot 10^5$ and $3.5 \cdot 10^5$ environment steps in order to converge to the exact responsibility assignment in these two configurations.
In comparison, performing an exhaustive search, i.e., fully executing BF-DT or BF-ST, on one trajectory of \textit{Euchre}$(10)$ and one of \textit{Spades}$(10)$ can take up to more than $10 \cdot 10^9$ and $6.2 \cdot 10^9$ steps, respectively.
It is also worth noting that RA-MCTS achieves $\epsilon_{max} = 0$ for more than $90\%$ of the runs, in the above mentioned configurations, within at most $2 \cdot 10^5$ and $0.5 \cdot 10^5$ steps.

By comparing RA-MCTS to RANDOM in Plots \ref{fig : no_unc_team_goofspiel_7}-\ref{fig : no_unc_spades_13}, we can see that the former always \textit{stochastically dominates} the latter \cite{levy1992stochastic}.\footnote{The curve of the dominant method is strictly above the other method's curve \cite{agarwal2021deep}.} 
As part of our ablation study, we also compare RA-MCTS to BF-ST. In Plots \ref{fig : no_unc_team_goofspiel_7}-\ref{fig : no_unc_team_goofspiel_9} and \ref{fig : no_unc_euchre_8}-\ref{fig : no_unc_spades_10}, RA-MCTS \textit{stochastically dominates} BF-ST. In Plot \ref{fig : no_unc_team_goofspiel_9}, performance profiles of the two methods are almost identical, while in Plot \ref{fig : no_unc_spades_13}, BF-ST outperforms RA-MCTS only for a small number of steps.
Moreover, it can be seen that almost always the maximum number of environment steps that RA-MCTS might need in order to find the exact responsibility assignment, is considerably less than that of BF-ST. For instance, in Plot \ref{fig : no_unc_team_goofspiel_8} RA-MCTS needs almost $5$ times fewer environment steps compared to BF-ST, while in Plots \ref{fig : no_unc_team_goofspiel_7}, \ref{fig : no_unc_euchre_8}, \ref{fig : no_unc_spades_9}, \ref{fig : no_unc_spades_10} we witness a drop of at least $30\%$. These results show that components from Sections \ref{sec.pruning} and \ref{sec.mcts} are important for RA-MCTS.
In \iftoggle{arxiv}{Appendix \ref{app.ablation}}{the Supplemental Material}, we include a similar ablation study, where we compare RA-MCTS to the BF-ST method enhanced with the pruning technique described in Section \ref{sec.pruning}.

Finally, Plots \ref{fig : no_unc_team_goofspiel_7}-\ref{fig : no_unc_spades_13} showcase that BF-ST \textit{stochastically dominates} BF-DT.
This result demonstrates that a brute force algorithm that uses the structure of the tree we propose in Section \ref{sec.pruning} converges faster to the exact solution than a brute force algorithm that uses the standard decision tree of the underlying Dec-POMDP.

\subsubsection{Results with Unknown Context}\label{sec.unc_results}

In this section, we present the results of our experiments under context uncertainty, where we make use of the sampling procedure introduced in Section \ref{sec.alg_unc}, with total number of samples $M=10$.
Each sample corresponds to a context generated by posterior inference.
Plots \ref{fig : unc_team_goofspiel_9}-\ref{fig : unc_spades_10} display performance profiles  for methods RA-MCTS and RANDOM for different computational budgets and for different values of threshold $D$.
First, we observe that almost always RA-MCTS stochastically dominates RANDOM.
Moreover, all plots show that RA-MCTS achieves $\epsilon_{max} = 0.25$ in more than $75\%$ of the runs, within at most $0.5 \cdot 10^5$ steps per sample, or $5 \cdot 10^5$ steps in total.
Finally, we can also see that our method performs the best for \textit{TeamGoofspiel}$(9)$, where it achieves $\epsilon_{max} = 0.15$ in $86\%$ of the runs.

Our results give prominence to an inherent problem of responsibility attribution under context uncertainty. We observe that even for a number of steps that suffices for RA-MCTS to find the exact agents' degrees of responsibility for most of the sampled trajectories, the responsibility assignments for many of the actual trajectories are not fully found.
We conclude then that even under unbounded computational budget if the posterior distribution of the underlying context of a trajectory is not informative enough, then failing to exactly estimate the agents' degrees of responsibility for that trajectory is unavoidable.
We believe however that one potential way to alleviate this issue is by designing responsibility attribution mechanisms that incorporate domain knowledge, which could balance the non-informativeness of the posterior distribution.


\section{Conclusion}\label{sec.conclusion}

We initiate the study of developing efficient algorithmic approaches for responsibility attribution in Dec-POMDPs. To that end, we propose and experimentally evaluate RA-MCTS, an MCTS type of method which efficiently approximates responsibility assignments.
Looking forward, we plan to apply and test the efficiency of RA-MCTS on a real-world domain. 
Extending our approach to continuous models is another research direction that we deem particularly interesting.

\section*{Acknowledgements}
This research was, in part, funded by the Deutsche Forschungsgemeinschaft (DFG, German Research Foundation) – project number $467367360$.



\bibliographystyle{ACM-Reference-Format} 
\bibliography{main}

\iftoggle{arxiv}{
\clearpage
\appendix


\section{List of Appendices}

In this section, we provide a brief description of the content provided in the appendices of the paper.

\begin{itemize}
    \item Appendix \ref{app.tlc} provides an extended discussion on the application scenario of autonomous traffic light control.
    \item Appendix \ref{app.gumbel} provides additional information on Gumbel-Max SCMs.
    \item Appendix \ref{app.euchre}  provides additional details on the players' policies in the \textit{Euchre} environment from Section \ref{sec.envs_policies}.
    \item Appendix \ref{app.spades} provides additional details on the players' policies in the \textit{Spades} environment from Section \ref{sec.envs_policies}.
    \item Appendix \ref{app.lower} provides the details of our experimental setup for large size environments.
    \item Appendix \ref{app.ablation} includes the results of an additional ablation study on RA-MCTS.
    \item Appendix \ref{app.unc} includes the remaining plots from Section \ref{sec.unc_results}. 
\end{itemize}


\section{Extended Discussion on Possible Application Scenario}\label{app.tlc}

In this section we discuss a possible application scenario for our proposed algorithmic framework. We consider autonomous traffic light control (ATLC) systems \cite{arel2010reinforcement}. Typically, in ATLC each agent controls one road intersection. At every time-step, the agent observes some local information, and depending on the system's exact implementation it might as well receive messages from other agents. Based on this information, the agent then decides on how to schedule the traffic lights of the intersection it controls. A common failure example in ATLC includes a driver waiting in some intersection for more than an ``acceptable'' time period. In systems as complex as ATLC, however, it is hard to identify who might have caused such a failure. This is due to the temporal dependencies between the agents' decisions. The system designers might want to determine, for instance, whether the agent who controls the intersection where the driver had to wait made a mistake? Or was the failure (partly) caused by mistakes made by other agents in previous time-steps?
This is where our algorithmic framework for attributing responsibility can prove useful. In a scenario like this, our method can efficiently approximate the degree to which each agent contributed to the system's failure.

From a technical perspective, we mention that our search method is generic and can be technically applied to any setting modeled as a finite and discrete Dec-POMDP. There are recent works that model ATLC as Dec-POMDPs \cite{jiang2022multi}.  


\section{Gumbel-Max SCMs}\label{app.gumbel}

In this section, we show how to implement Gumbel-Max SCMs in the Dec-POMDP SCM setting. Furthermore, we provide intuition behind a desirable property they satisfy.
For more details, we refer the interested reader to \cite{oberst2019counterfactual}.

Under the Gumbel-Max model, the structural equations of Eq. \eqref{eq.struct_eq} become:
\begin{align*}\label{eq.gum_struct_eq}
    & S_t = \argmax_{s \in \mathcal{S}}\{\log P(S_{t-1}, A_{t-1}, S_t = s) + U_{S_t}\}\\
    & O_t = \argmax_{o \in \mathcal{O}}\{\log \Omega(S_t, O_t = o) + U_{O_t}\}\\
    & I_{i,t} = \argmax_{\imath_i \in \mathcal{I}_i}\{\log Z_i(I_{i,t-1}, A_{i,t-1}, O_t, I_{i,t} = \imath_i) + U_{I_{i,t}}\}\\
    & A_{i,t} = \argmax_{a_i \in \mathcal{A}_i}\{\log \pi_i(A_{i, t} = a_i|I_{i,t}) + U_{A_{i,t}}\}
\end{align*}
where  $U_{S_t}$, $U_{O_t}$, $U_{I_{i,t}}$ and $U_{A_{i,t}} \sim$ Gumbel$(0, 1)$. 

It has been shown that the class of Gumbel-Max SCMs satisfies a
desirable property for categorical SCMs, namely the counterfactual
stability property. This property is considered important because it excludes a specific type of non-intuitive counterfactual outcomes. What follows, is an example taken from \cite{10.1145/3514094.3534133} which aims to provide the main intuition behind this property.
Consider the observed trajectory $\tau = \{(s_t, a_t)\}_{t=0}^{T-1}$, and the counterfactual scenario in which agents $\{1, ..., n\}$ take the joint action $a'$ at time-step $t$, instead of $a_t$. The counterfactual stability property ensures that under this counterfactual scenario, it is impossible that at time-step $t+1$ the process would transition to a state $s'$ different than the observed state, i.e., $s_{t+1}$, if the following condition holds
\begin{align*}
    \frac{P(S_t = s, A_t = a', S_{t+1} = s_{t+1})}{P(S_t = s, A_t = a_t, S_{t+1} = s_{t+1})} \ge 
    \frac{P(S_t = s, A_t = a', S_{t+1} = s')}{P(S_t = s, A_t = a_t, S_{t+1} = s')}.
\end{align*}
In other words, in order for the state at time-step $t+1$ to change under a counterfactual scenario, the relative likelihood of an alternative state $s'$ must have increased compared to that of $s_{t+1}$.


\section{Additional Details on Euchre}\label{app.euchre}

In this section, we present the agents' and opponents' for the card game \textit{Euchre}, which was described in Section \ref{sec.envs_policies}.

\textbf{Information states} of all players include the \textit{trump suit}, the \textit{leading suit} of the current round, their hand and the current trick.

\textbf{Agents} have deterministic policies which differ from each other's. If they are the first to play on a round: \agentzero~plays the lowest ranked card on its hand, while \agentone~plays its highest ranked card that (if possible) is not from the trump suit. 
When playing second or third on a round: both agents play their worst valid card if their teammate is leading the round or if they do not have any winning card, otherwise \agentzero~plays the worst of its winning cards and \agentone~plays its best such card.
When playing last: they both play their lowest ranked winning card, and if they do not have a winning card then they just play their lowest ranked valid card.

\textbf{Opponents} share the same stochastic policy.
If they are the first to play on a round: then they pick a card from their hand uniformly at random.
When playing second or third on a round: they play with $0.8$ probability a winning card or their lowest ranked valid card in case they do not have winning cards.
When playing last: they play with $0.8$ probability a winning card or their lowest ranked valid card in case they do not have winning cards or their teammate is leading the round.


\section{Additional Details on Spades}\label{app.spades}

In this section, we present the agents' and opponents' policies for the card game \textit{Spades}, which was described in Section \ref{sec.envs_policies}. Each player's policy is divided into a bidding policy and playing policy. The former is activated during the bidding phase of the game, while the latter in the playing phase.
The playing policies of the players are almost identical to the ones the players follow in \textit{Euchre}, and can be found in Appendix \ref{app.euchre}.
Regarding their bidding policies, all players adhere to the following mechanism which is based on the player's initial hand.
The player starts with bid $0$, and increases it by $1$ for every for every \textit{King} and \textit{Ace} they have, and also if they have the \textit{Jack} or \textit{Queen} of spades. They also add to their bid the total number of cards they have with suit of spades and value below \textit{Jack}, divided by $\frac{H}{3}$, where $H$ is the initial size of their hand.


\section{Lower Bounds of Responsibility Degrees for Large Environments}\label{app.lower}

In this section, we describe the procedure that we follow to generate a trajectory for which we can compute lower bounds of the agents’ degrees of responsibility, when the size of the environment is too large to compute them exactly (see Section \ref{sec.setup}). Note that the same procedure is used for all the environments.

First, we randomly pick $5$ action variables for each agent. For each one of those variables, we ``poison'' the agent's policy, such that zero probability is assigned to the action that the agent's non-poisoned (deterministic) policy would have normally taken, and all other valid actions are assigned equal probabilities.\footnote{We have also implemented the poisoning procedure for stochastic policies.}
Next, we sample a trajectory in which the agents win by using their non-poisoned policies.
We then fix the context that was used to generate the previous trajectory, and simulate the agents' poisoned policies in the environment.
We repeat the last two steps, until the outcome of the latter trajectory signifies a loss for the agents.
In order to compute lower bounds for the agents' degrees of responsibility on the failed trajectory, all we have to do is execute a brute force method (e.g., BF-DT) restricted to the action variables that were poisoned.

Note that there can be actual cause-witness pairs with variables that are not checked by the search, and hence why the above mentioned procedure provides lower bounds of the responsibilities and not their exact values. The pool of action variables the brute force is restricted to however, is promising for finding actual cause-witness pairs, and that is why the lower bounds this mechanism computes are not trivial.


\section{Additional Ablation Study}\label{app.ablation}

\captionsetup[figure]{belowskip=-10pt}
\begin{figure*}
\captionsetup[subfigure]{aboveskip=0pt,belowskip=1pt}
\centering
    \begin{subfigure}[c]{0.32\textwidth}
        \includegraphics[width=\textwidth]{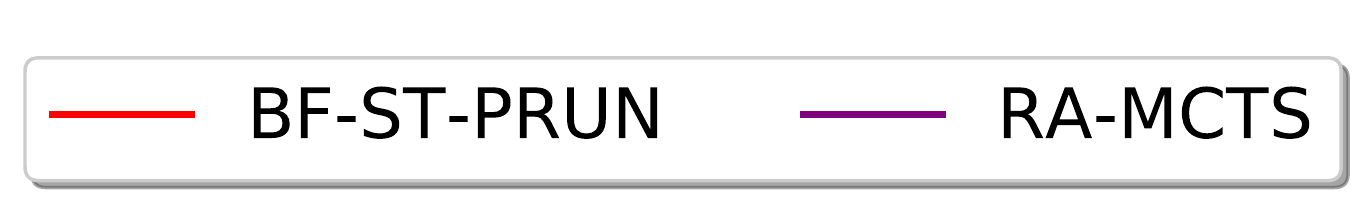}
    \end{subfigure}\\
    \begin{subfigure}[c]{0.24\textwidth}
        \includegraphics[width=\textwidth]{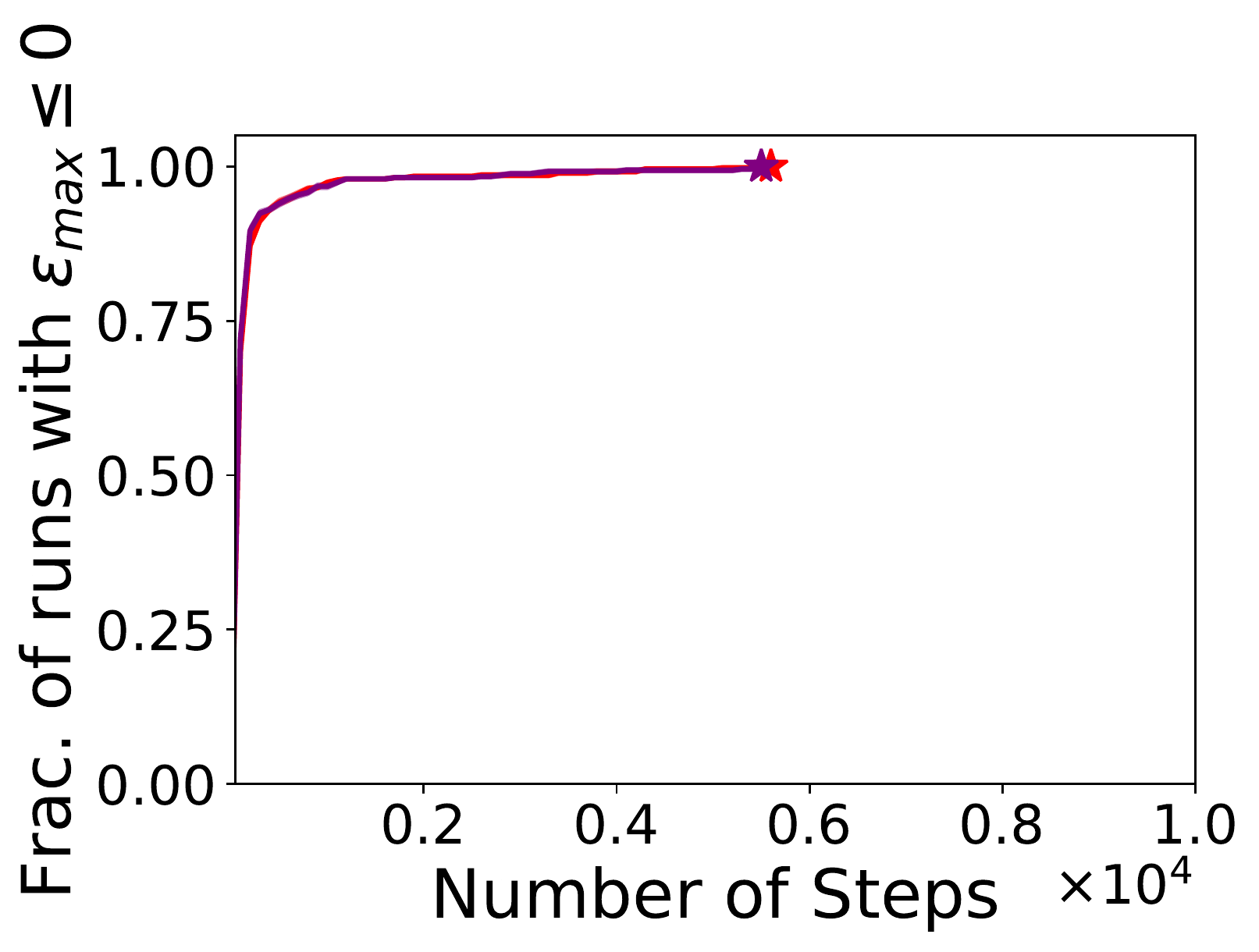}
        \captionsetup{type=figure}
        \caption{\textit{TeamGoofspiel}$\boldsymbol{(7)}$}
        \label{fig : ablation_team_goofspiel_7}
    \end{subfigure}\hfill%
    \begin{subfigure}[c]{0.24\textwidth}
        \includegraphics[width=\textwidth]{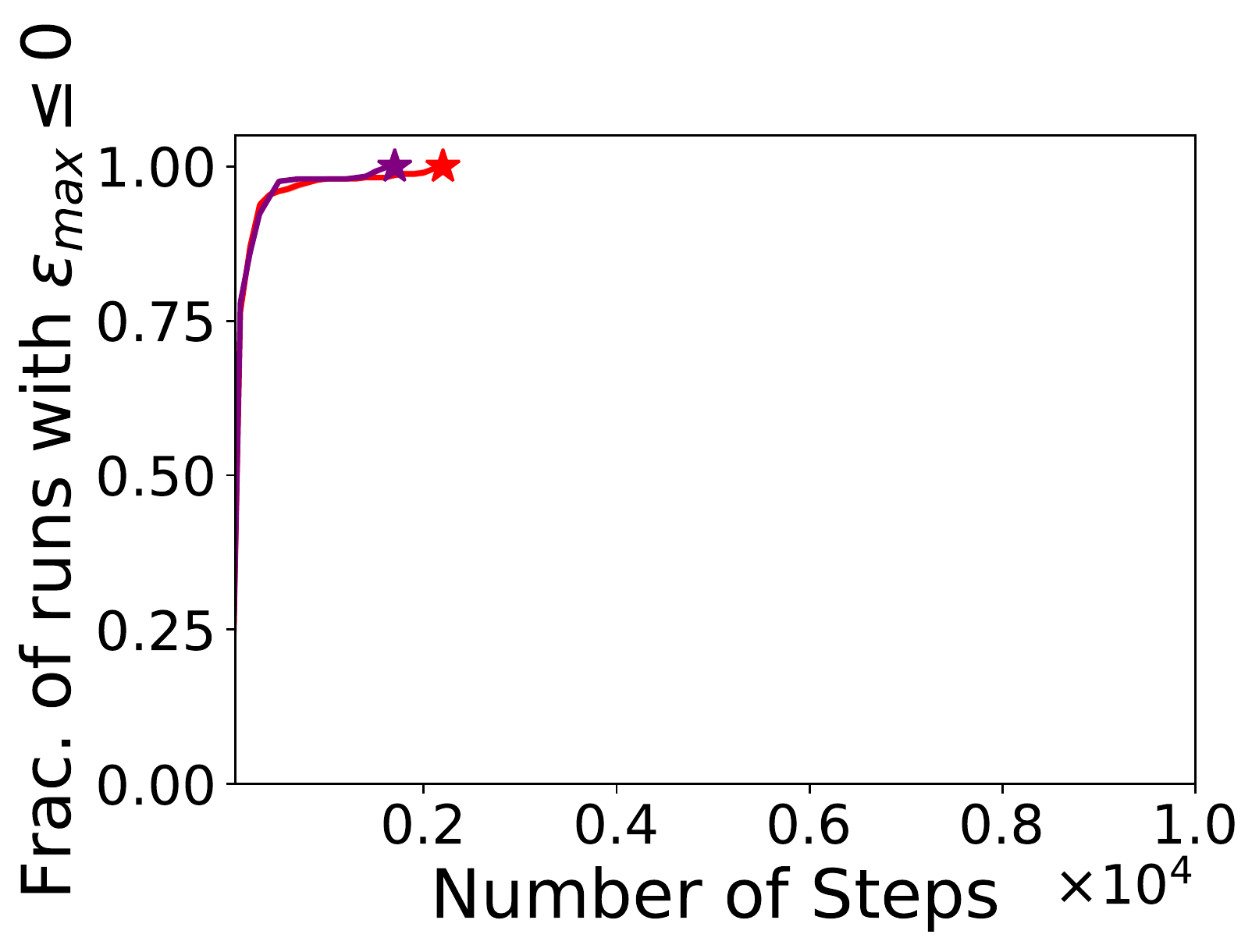}
        \captionsetup{type=figure}
        \caption{\textit{TeamGoofspiel}$\boldsymbol{(8)}$}
        \label{fig : ablation_team_goofspiel_8}
    \end{subfigure}\hfill%
    \begin{subfigure}[c]{0.24\textwidth}
        \includegraphics[width=\textwidth]{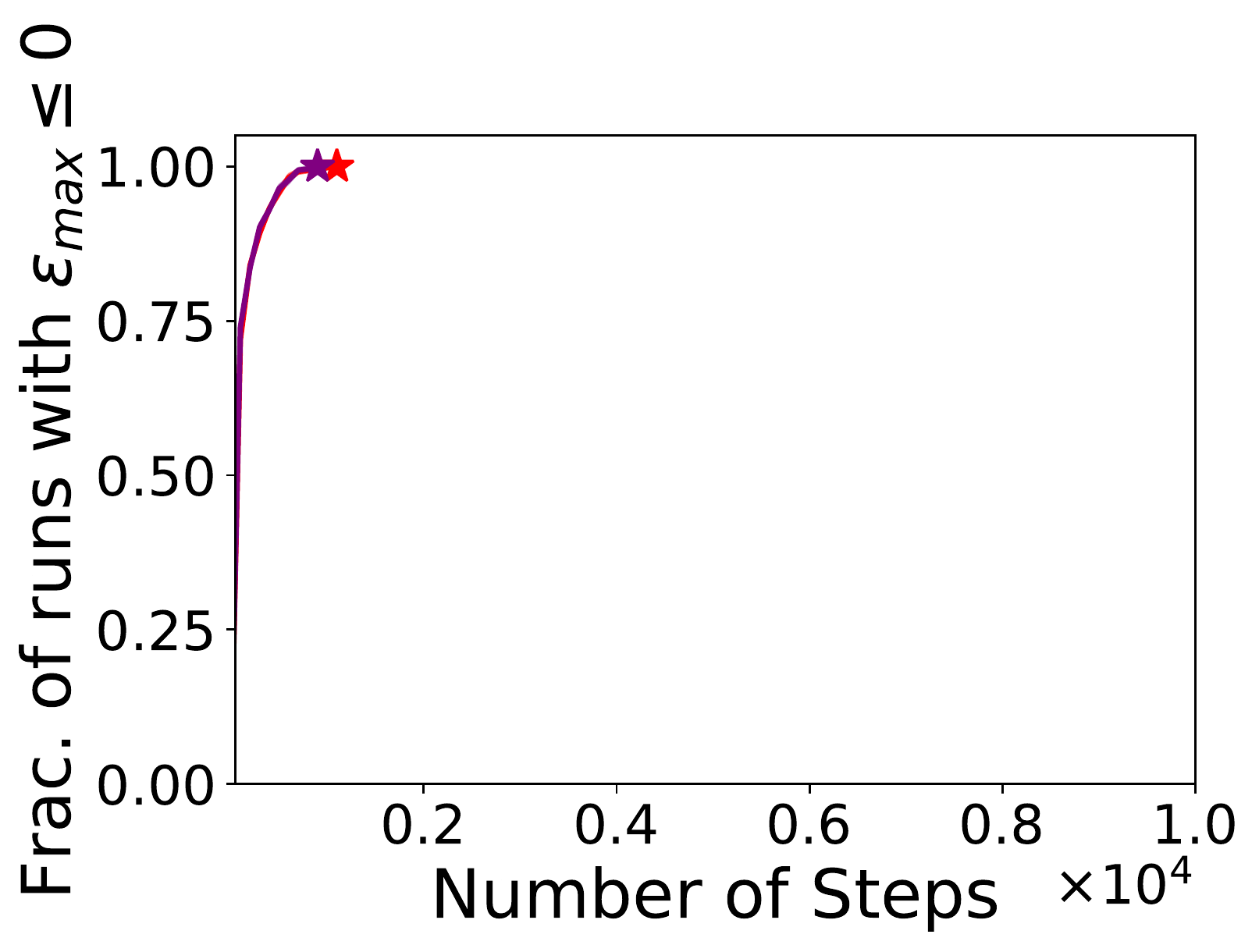}
        \captionsetup{type=figure}
        \caption{\textit{TeamGoofspiel}$\boldsymbol{(9)}$}
        \label{fig : ablation_team_goofspiel_9}
    \end{subfigure}
    \begin{subfigure}[c]{0.24\textwidth}
        \includegraphics[width=\textwidth]{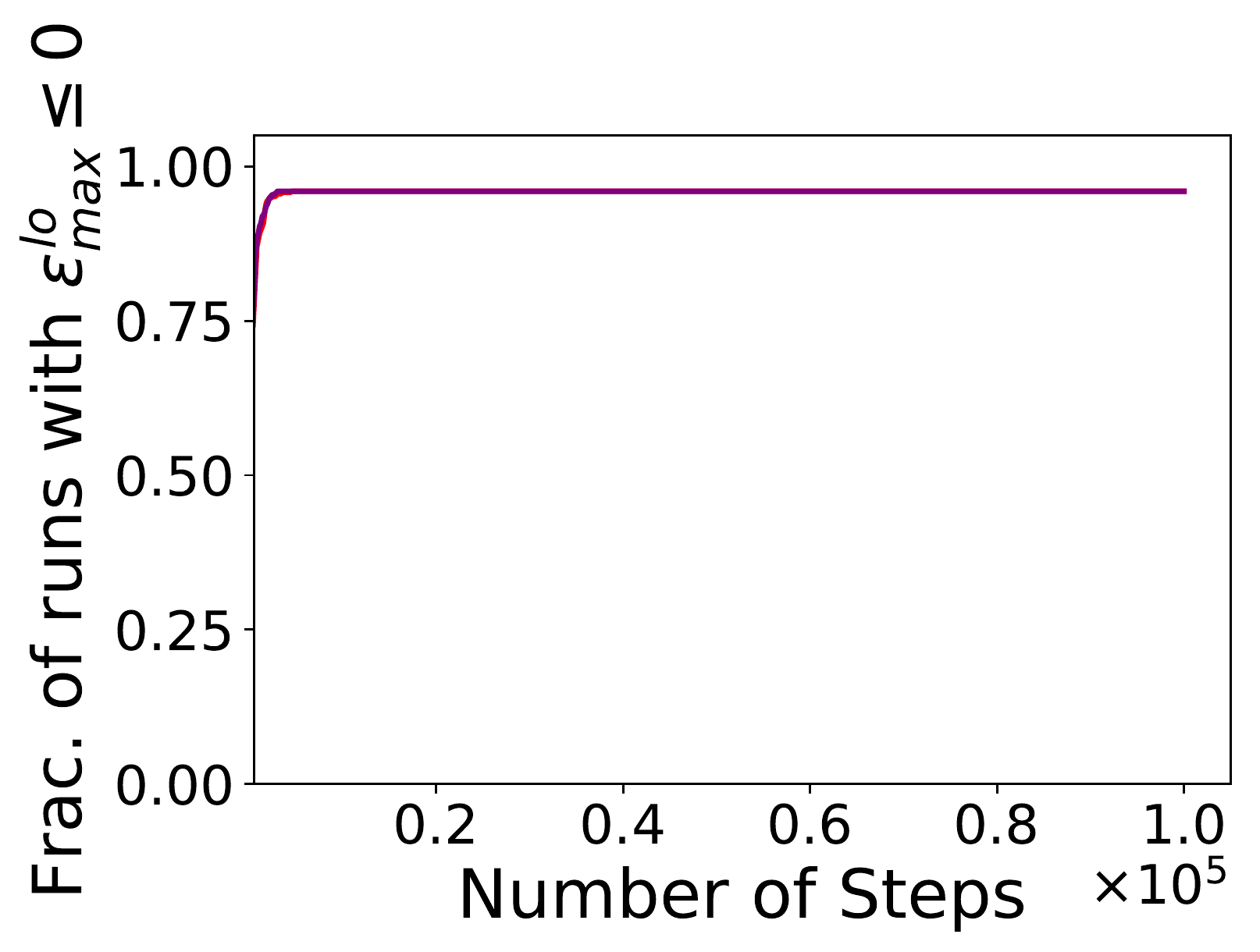}
        \captionsetup{type=figure}
        \caption{\textit{TeamGoofspiel}$\boldsymbol{(13)}$}
        \label{fig : ablation_team_goofspiel_13}
    \end{subfigure}\\
    \begin{subfigure}[c]{0.24\textwidth}
        \includegraphics[width=\textwidth]{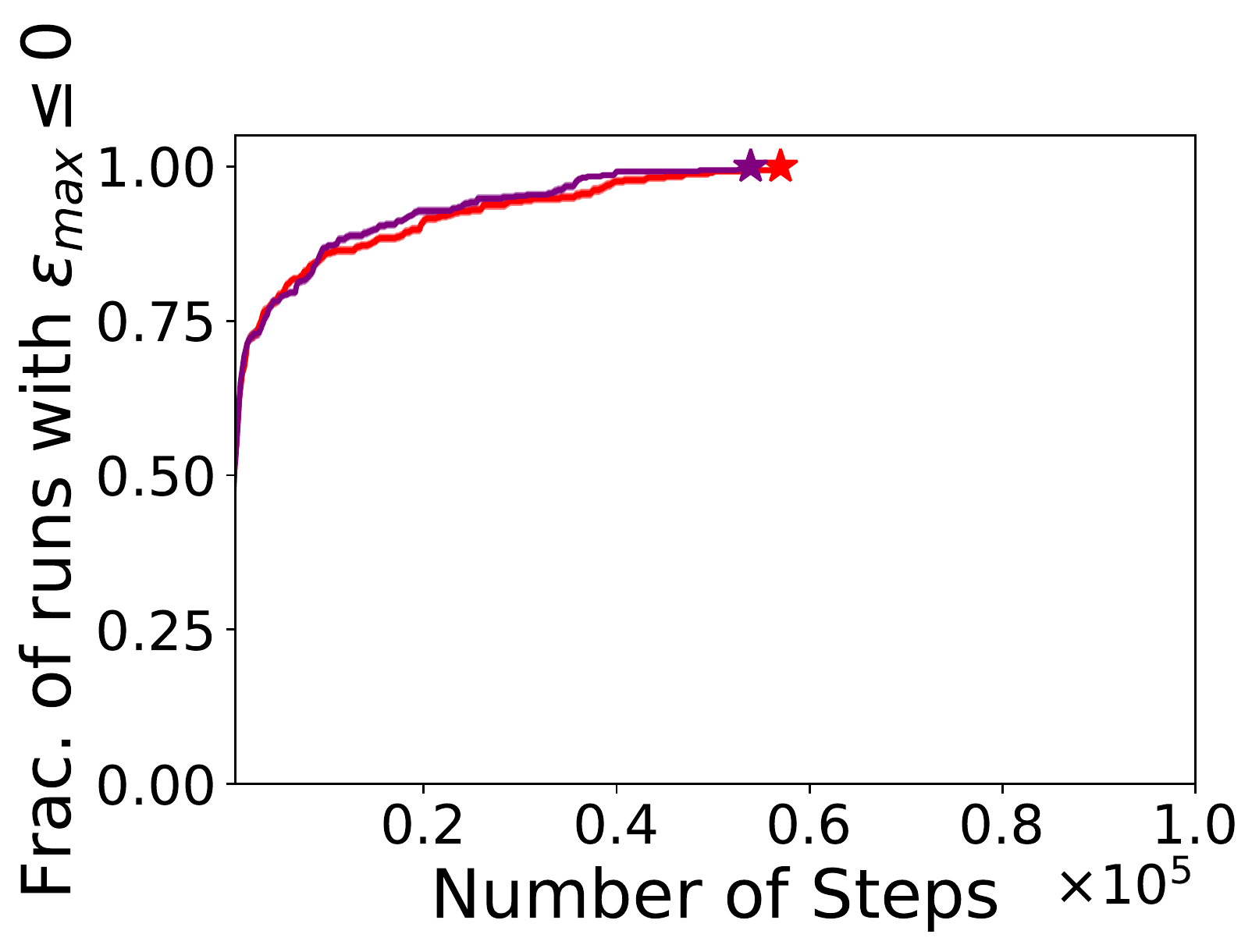}
        \captionsetup{type=figure}
        \caption{\textit{Euchre}$\boldsymbol{(8)}$}
        \label{fig : ablation_euchre_8}
    \end{subfigure}\hfill%
    \begin{subfigure}[c]{0.24\textwidth}
        \includegraphics[width=\textwidth]{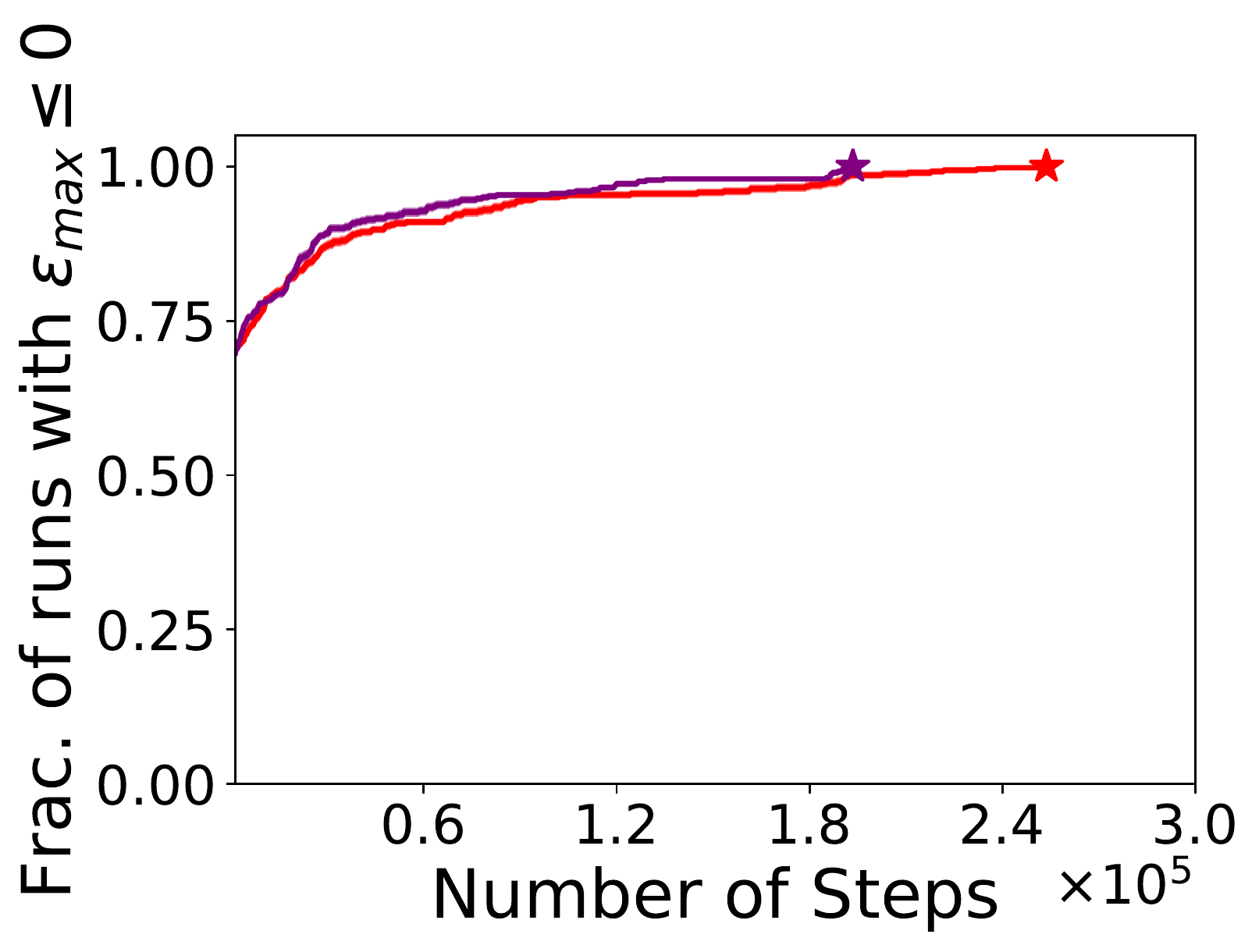}
        \captionsetup{type=figure}
        \caption{\textit{Euchre}$\boldsymbol{(9)}$}
        \label{fig : ablation_euchre_9}
    \end{subfigure}\hfill%
    \begin{subfigure}[c]{0.24\textwidth}
        \includegraphics[width=\textwidth]{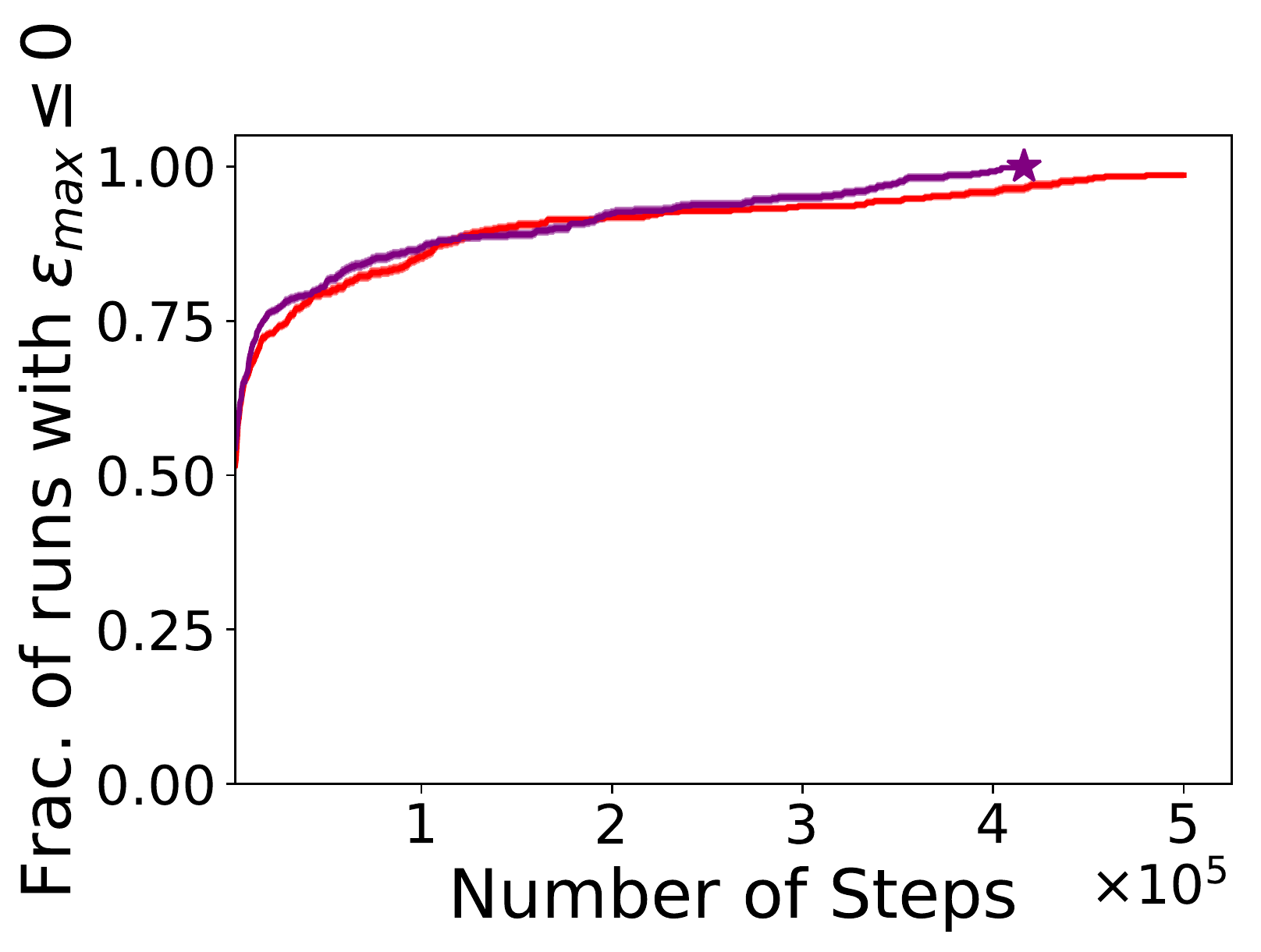}
        \captionsetup{type=figure}
        \caption{\textit{Euchre}$\boldsymbol{(10)}$}
        \label{fig : ablation_euchre_10}
    \end{subfigure}\hfill%
    \begin{subfigure}[c]{0.24\textwidth}
        \includegraphics[width=\textwidth]{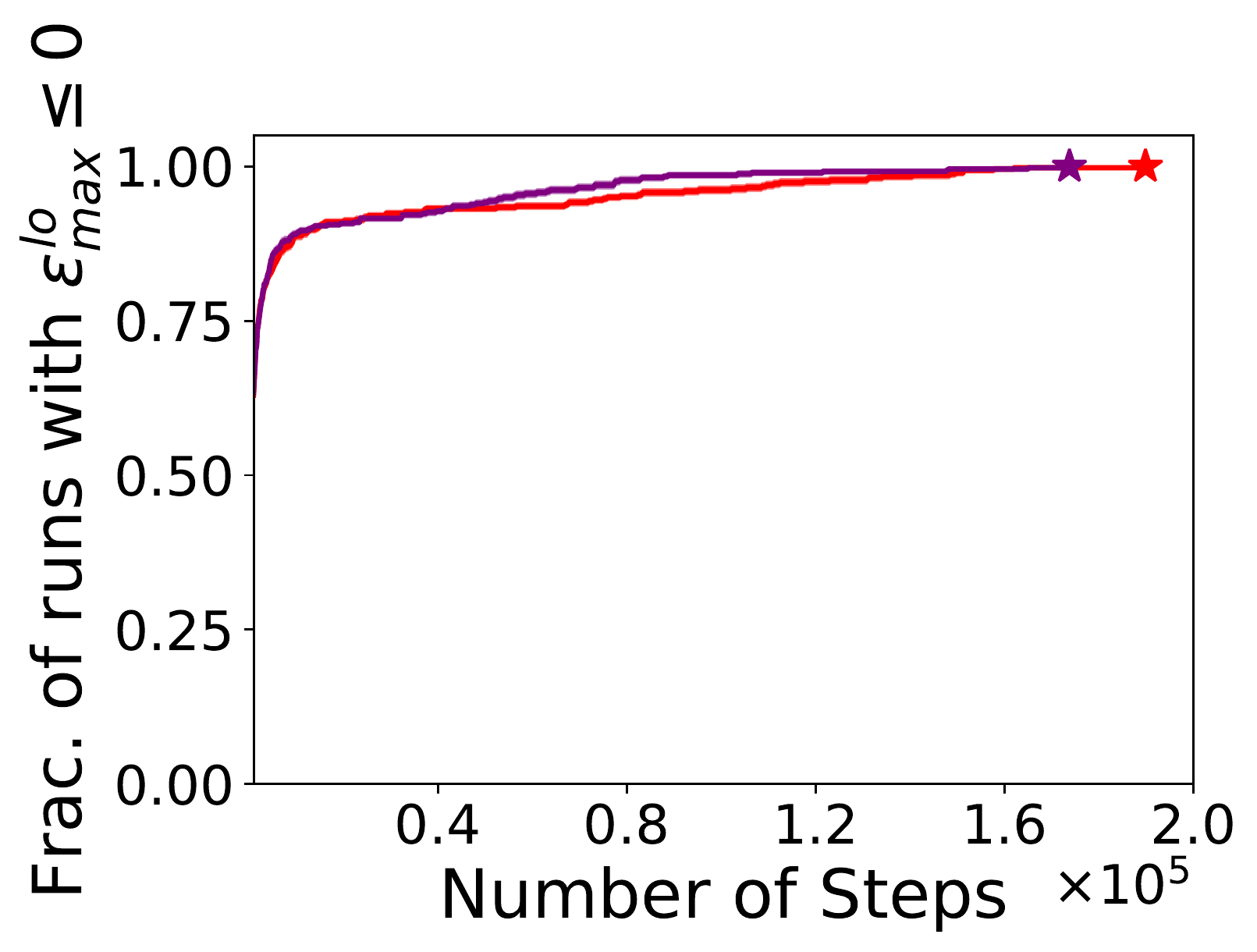}
        \captionsetup{type=figure}
        \caption{\textit{Euchre}$\boldsymbol{(12)}$}
        \label{fig : ablation_euchre_12}
    \end{subfigure}\\
    \begin{subfigure}[c]{0.24\textwidth}
        \includegraphics[width=\textwidth]{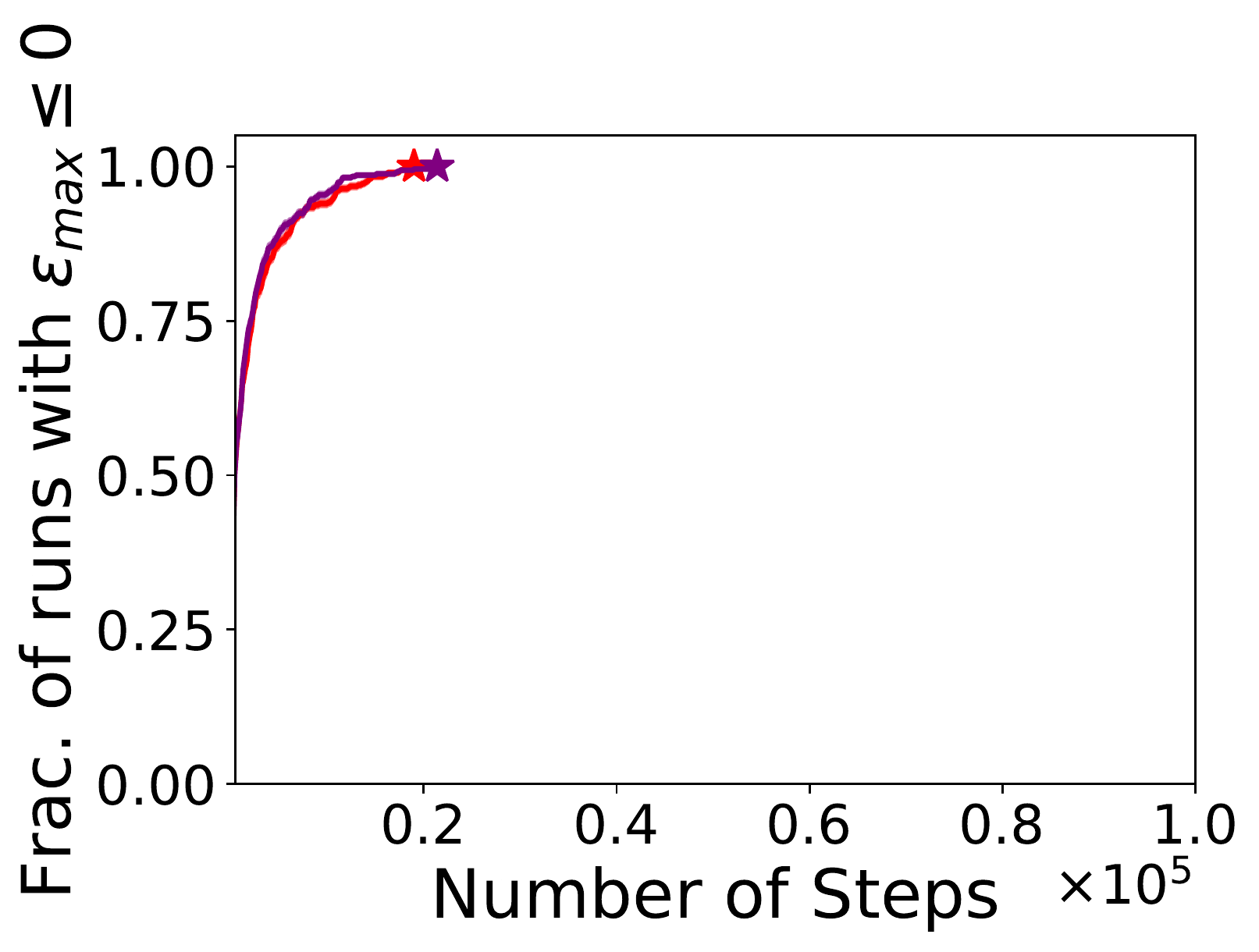}
        \captionsetup{type=figure}
        \caption{\textit{Spades}$\boldsymbol{(8)}$}
        \label{fig : ablation_spades_8}
    \end{subfigure}\hfill%
    \begin{subfigure}[c]{0.24\textwidth}
        \includegraphics[width=\textwidth]{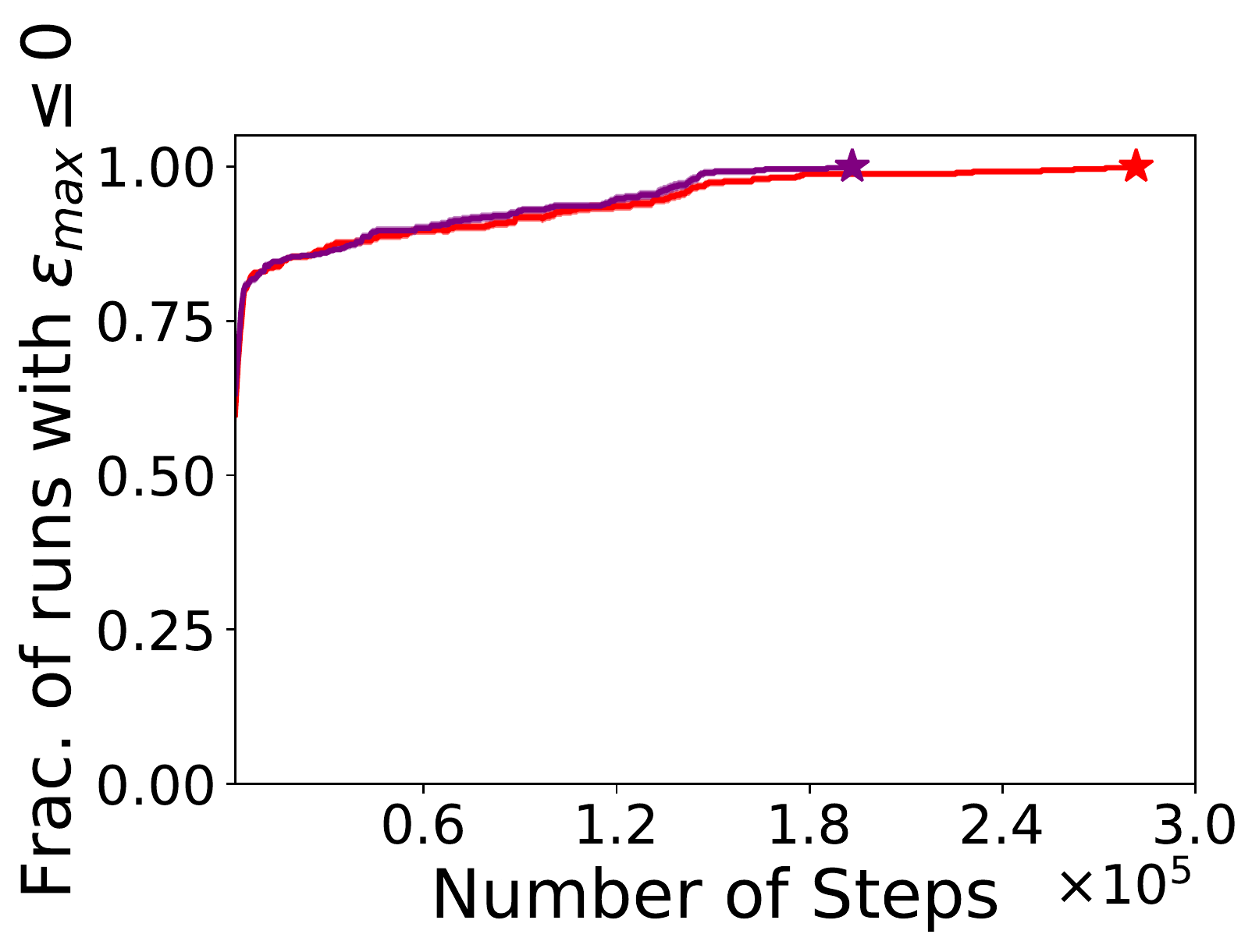}
        \captionsetup{type=figure}
        \caption{\textit{Spades}$\boldsymbol{(9)}$}
        \label{fig : ablation_spades_9}
    \end{subfigure}\hfill%
    \begin{subfigure}[c]{0.24\textwidth}
        \includegraphics[width=\textwidth]{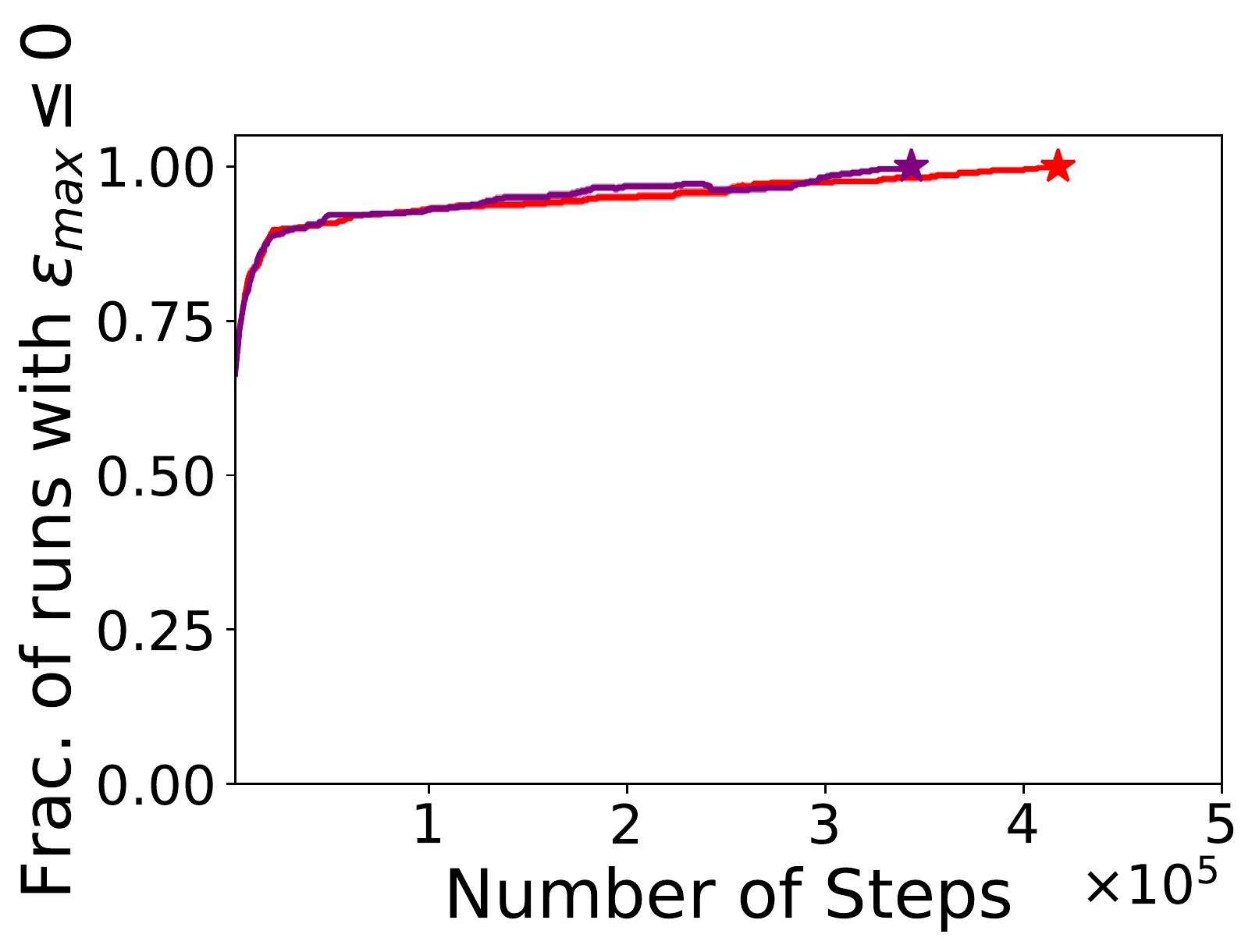}
        \captionsetup{type=figure}
        \caption{\textit{Spades}$\boldsymbol{(10)}$}
        \label{fig : ablation_spades_10}
    \end{subfigure}\hfill%
    \begin{subfigure}[c]{0.24\textwidth}
        \includegraphics[width=\textwidth]{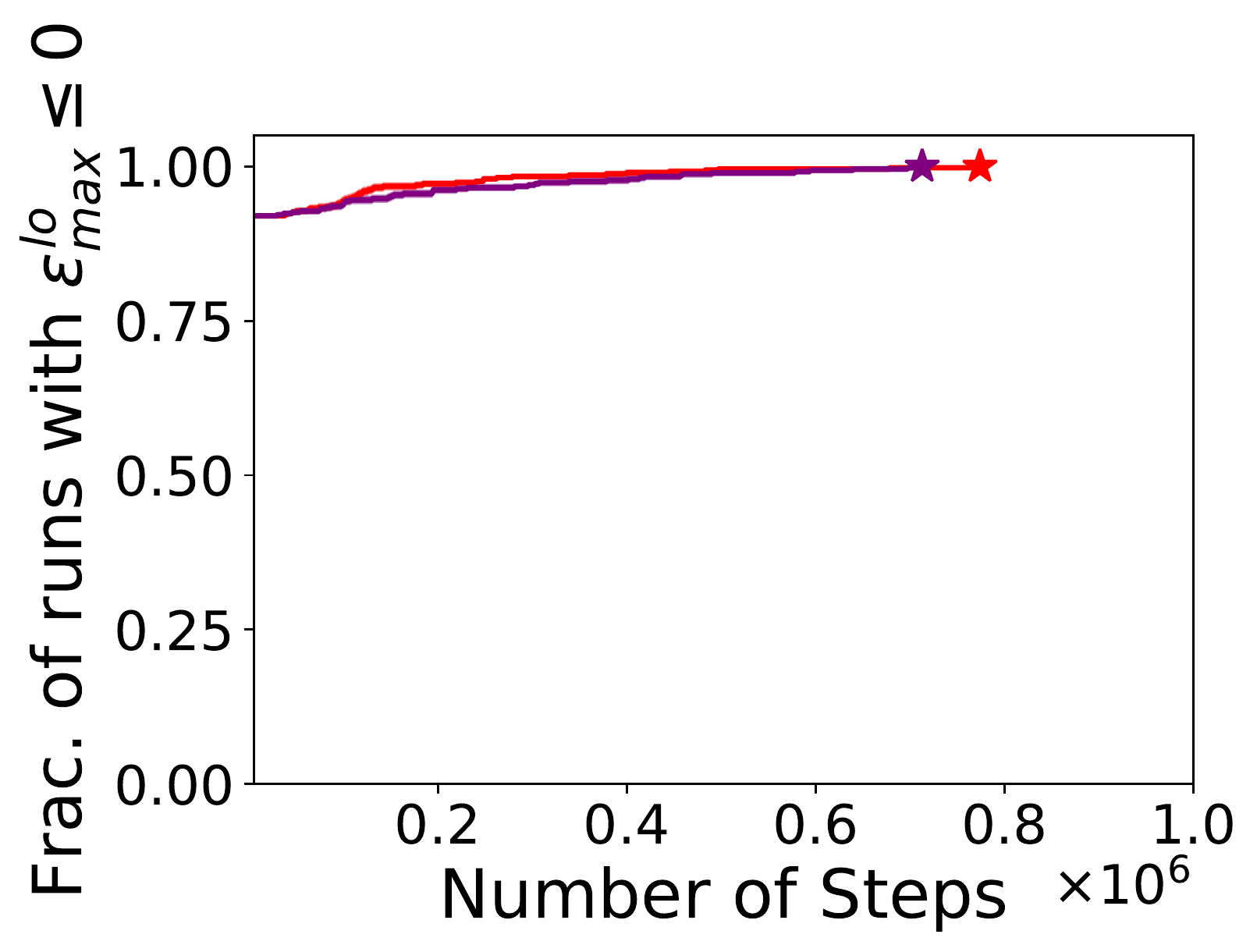}
        \captionsetup{type=figure}
        \caption{\textit{Spades}$\boldsymbol{(13)}$}
        \label{fig : ablation_spades_13}
    \end{subfigure}
\caption{Performance profiles on \textit{TeamGoofspiel}, \textit{Euchre} and \textit{Spades} with known context. Shaded regions show standard deviation. We add a marker if at the current number of steps the fraction of runs with $\epsilon_{max} \leq 0$ or $\epsilon^{lo}_{max} \leq 0$ is $1$. Plots in this figure are part of our ablation study, and they are used to compare RA-MCTS to BF-ST-PRUN.} 
\label{fig: ablation}
\end{figure*}

As part of our ablation study, we compare RA-MCTS to method BF-ST-PRUN, which is essentially the method BF-ST from Section \ref{sec.setup} enhanced with the pruning technique described in Section \ref{sec.pruning}. By looking at the plots in Fig. \ref{fig: ablation}, we conclude that the effects that can be seen when comparing RA-MCTS and BF-ST-PRUN, are similar albeit less prominent to those for RA-MCTS and BF-ST.
These results showcase that components from Section \ref{sec.mcts} are important for RA-MCTS.


\section{Additional Results with Unknown Context}\label{app.unc}

Fig. \ref{fig: app_unc_plots} contains the full set of plots for results with unknown context (Section \ref{sec.unc_results}).

\captionsetup[figure]{belowskip=-10pt}
\begin{figure*}
\captionsetup[subfigure]{aboveskip=0pt,belowskip=1pt}
\centering
    \begin{subfigure}[c]{\textwidth}
        \includegraphics[width=\textwidth]{./figures/unc/legend.pdf}
    \end{subfigure}\\
    \begin{subfigure}[c]{0.24\textwidth}
        \includegraphics[width=\textwidth]{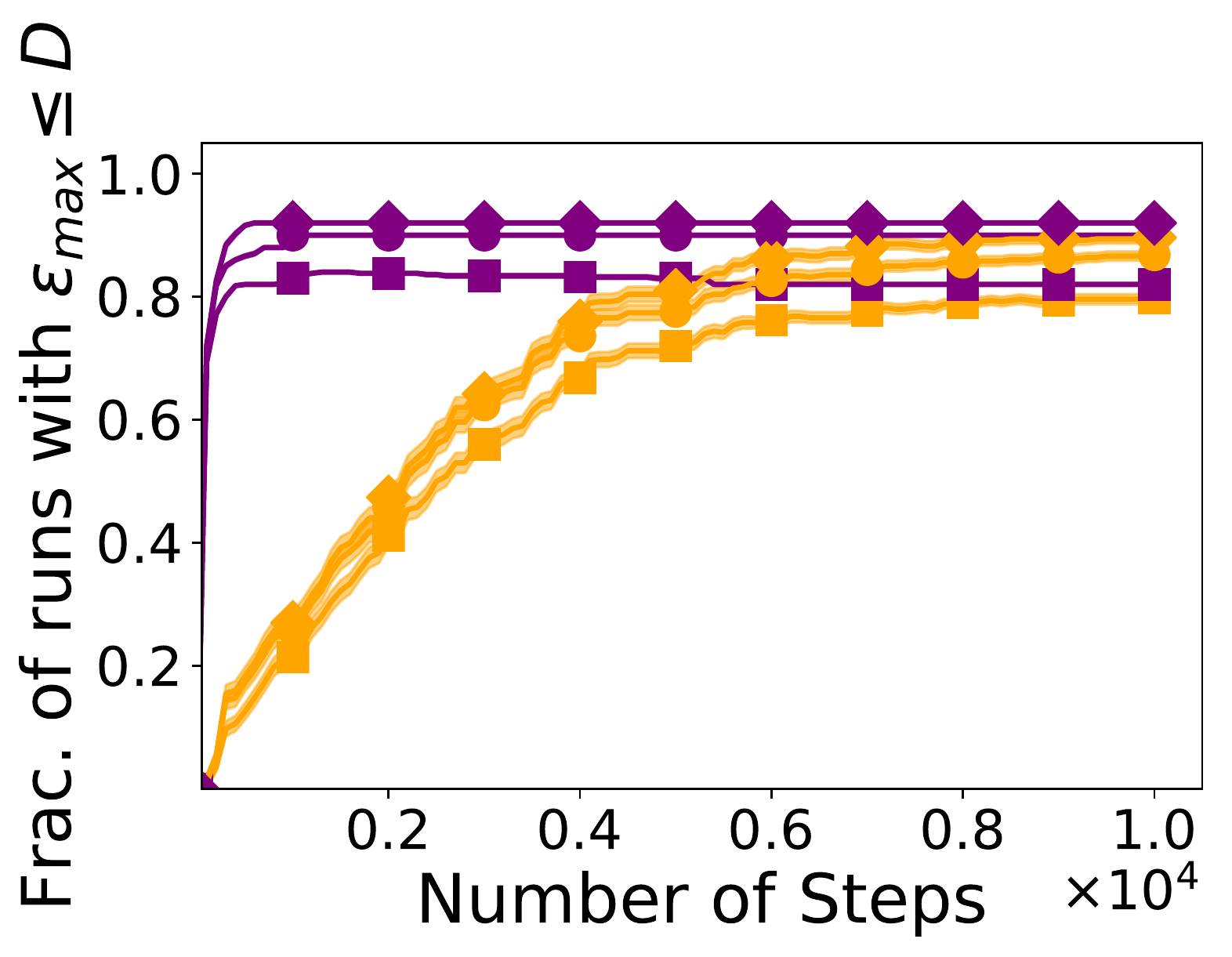}
        \captionsetup{type=figure}
        \caption{\textit{TeamGoofspiel}$\boldsymbol{(7)}$ Unc.}
        \label{fig : unc_team_goofspiel_7}
    \end{subfigure}\hfill%
    \begin{subfigure}[c]{0.24\textwidth}
        \includegraphics[width=\textwidth]{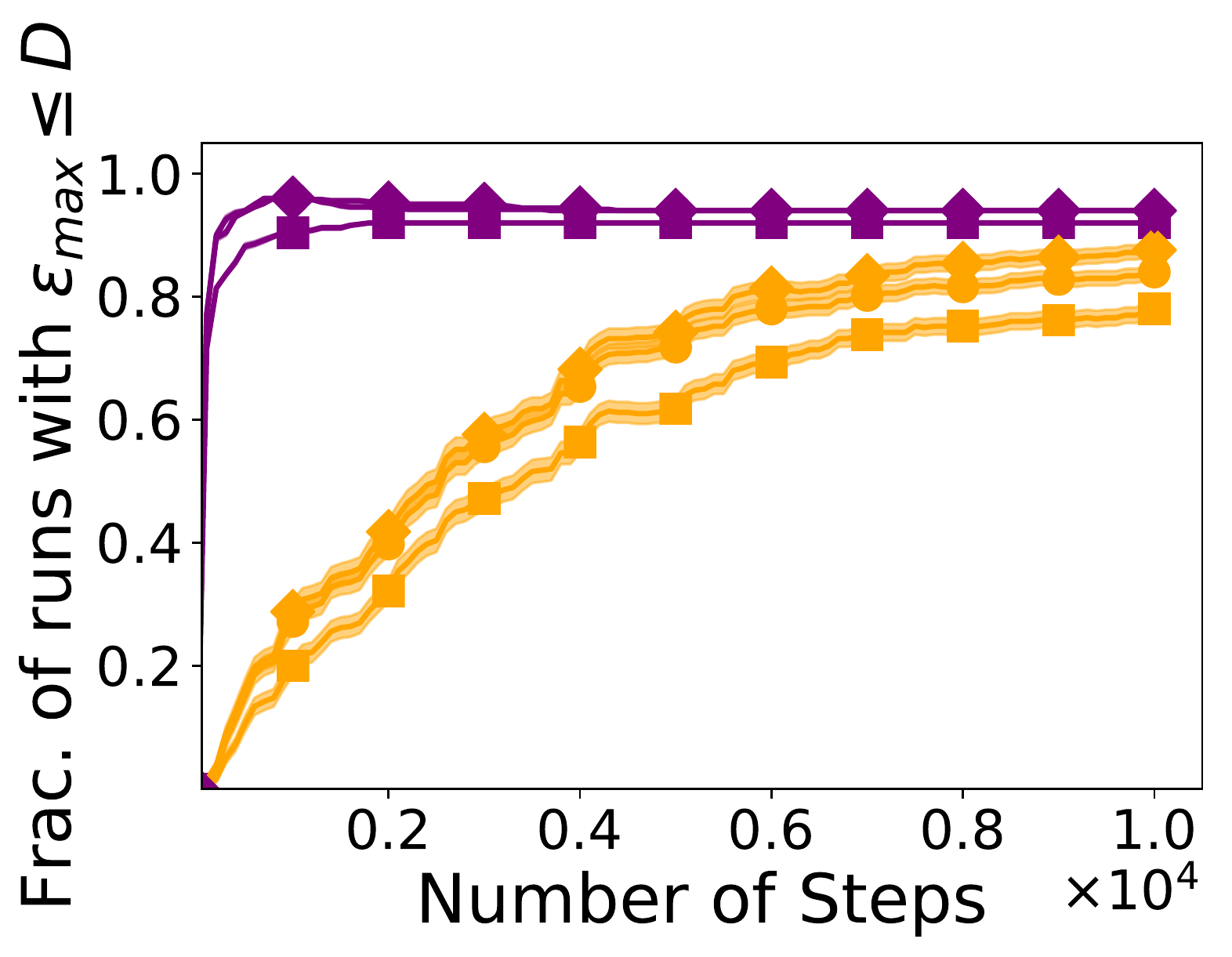}
        \captionsetup{type=figure}
        \caption{\textit{TeamGoofspiel}$\boldsymbol{(8)}$ Unc.}
        \label{fig : unc_team_goofspiel_8}
    \end{subfigure}\hfill%
    \begin{subfigure}[c]{0.24\textwidth}
        \includegraphics[width=\textwidth]{./figures/unc/team_goofspiel/env_size=9.pdf}
        \captionsetup{type=figure}
        \caption{\textit{TeamGoofspiel}$\boldsymbol{(9)}$ Unc.}
    \end{subfigure}
    \begin{subfigure}[c]{0.24\textwidth}
        \includegraphics[width=\textwidth]{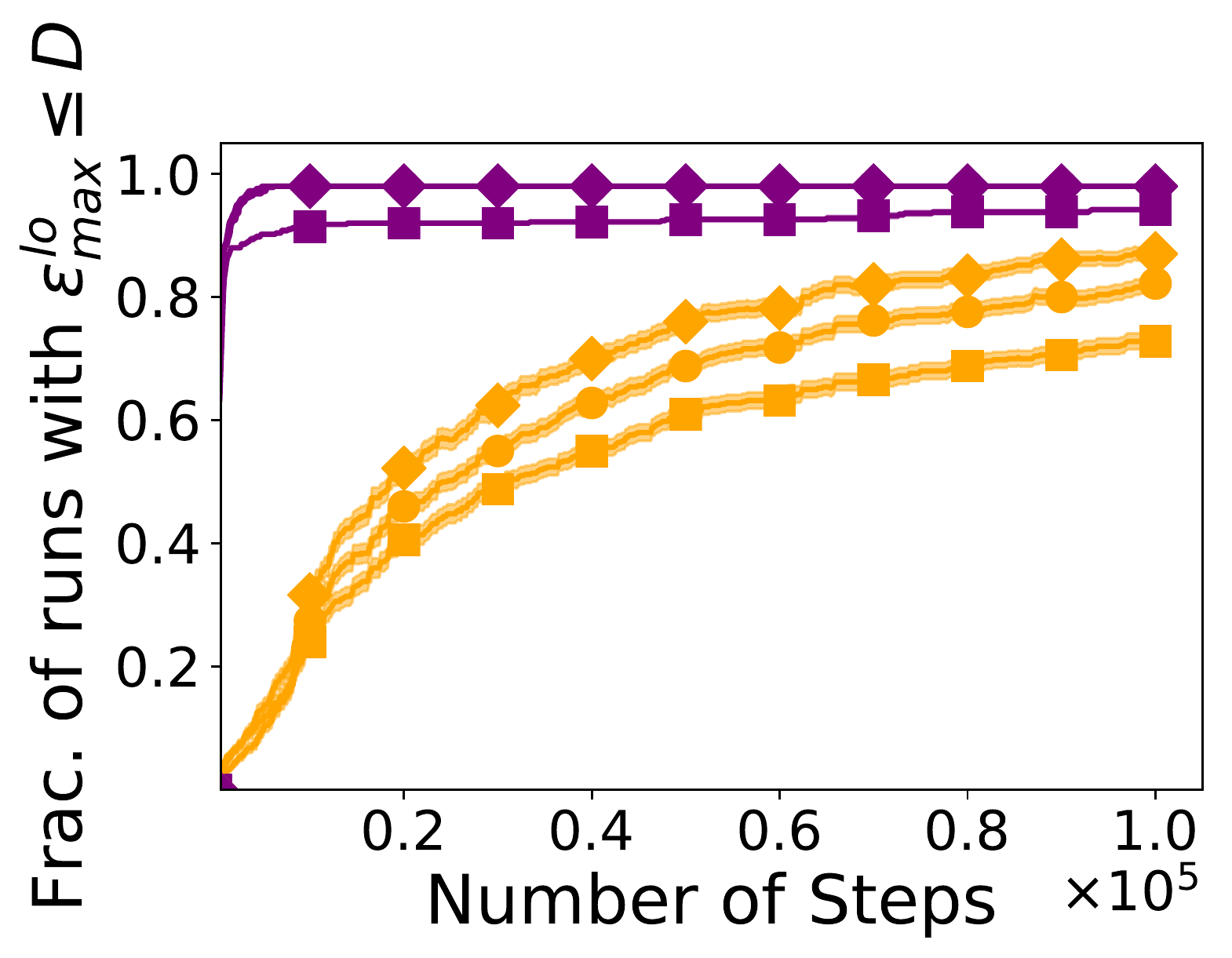}
        \captionsetup{type=figure}
        \caption{\textit{TeamGoofspiel}$\boldsymbol{(13)}$ Unc.}
        \label{fig : unc_team_goofspiel_13}
    \end{subfigure}\\
    \begin{subfigure}[c]{0.24\textwidth}
        \includegraphics[width=\textwidth]{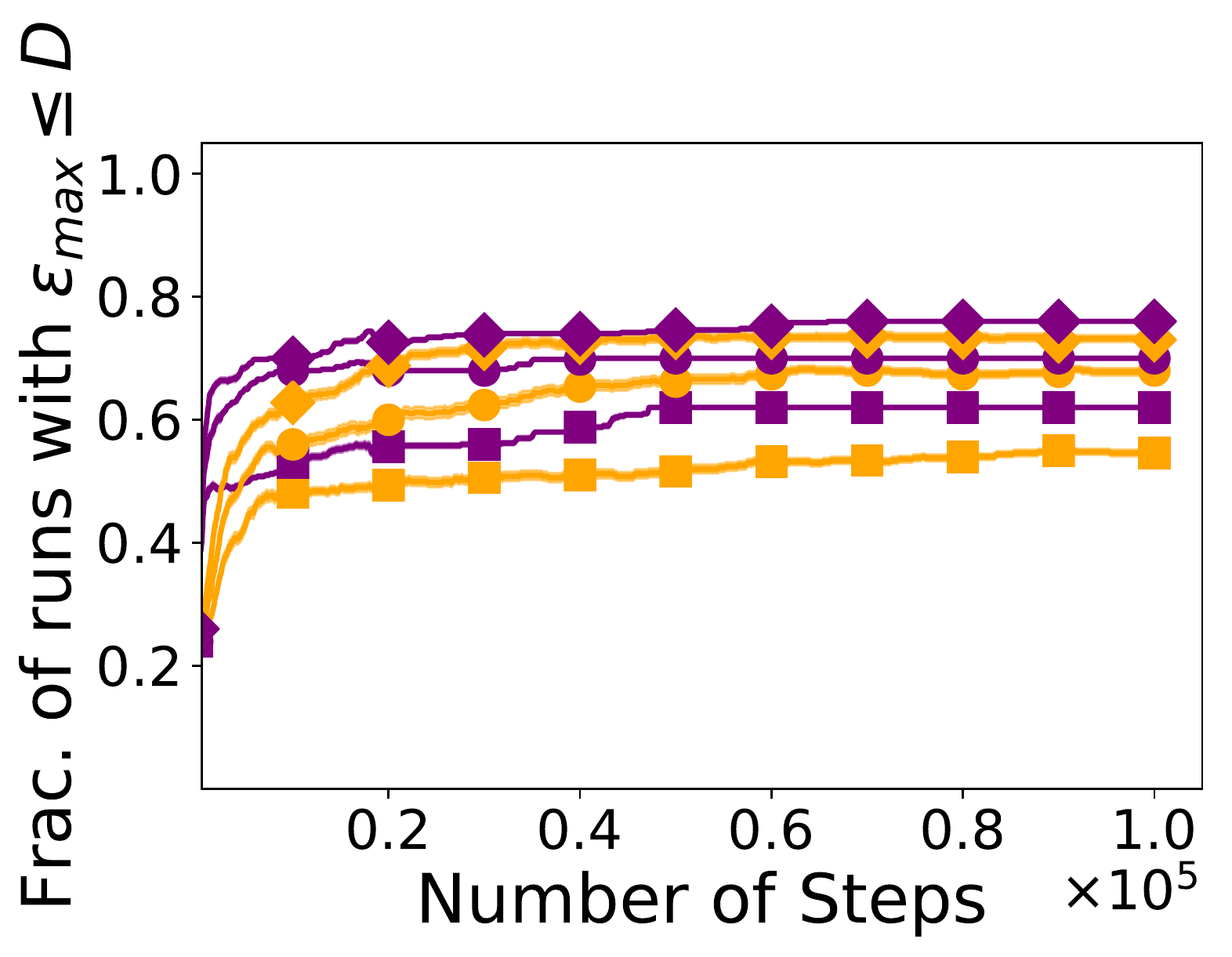}
        \captionsetup{type=figure}
        \caption{\textit{Euchre}$\boldsymbol{(8)}$ Unc.}
        \label{fig : unc_euchre_8}
    \end{subfigure}\hfill%
    \begin{subfigure}[c]{0.24\textwidth}
        \includegraphics[width=\textwidth]{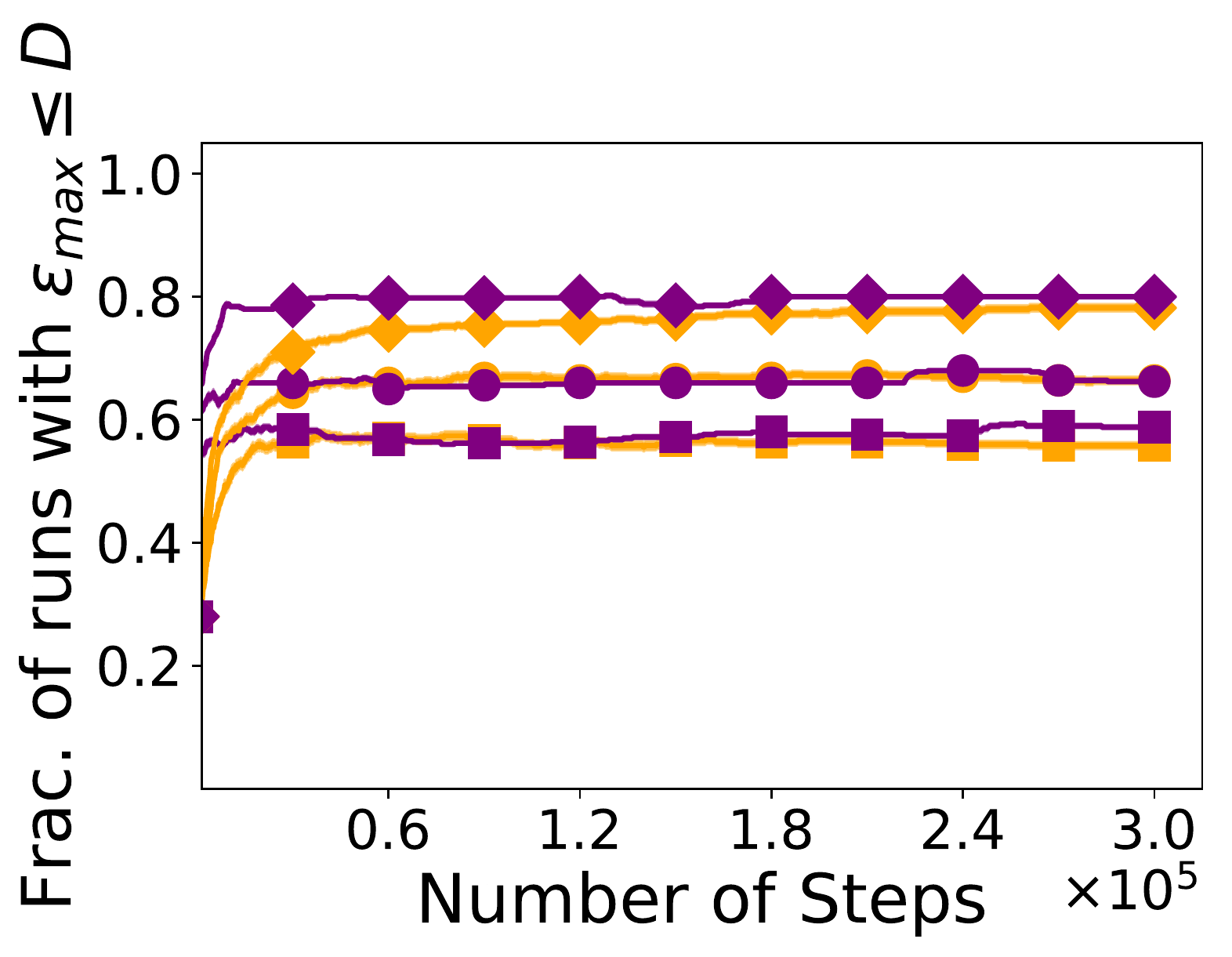}
        \captionsetup{type=figure}
        \caption{\textit{Euchre}$\boldsymbol{(9)}$ Unc.}
        \label{fig : unc_euchre_9}
    \end{subfigure}\hfill%
    \begin{subfigure}[c]{0.24\textwidth}
        \includegraphics[width=\textwidth]{./figures/unc/euchre/env_size=10.pdf}
        \captionsetup{type=figure}
        \caption{\textit{Euchre}$\boldsymbol{(10)}$ Unc.}
    \end{subfigure}\hfill%
    \begin{subfigure}[c]{0.24\textwidth}
        \includegraphics[width=\textwidth]{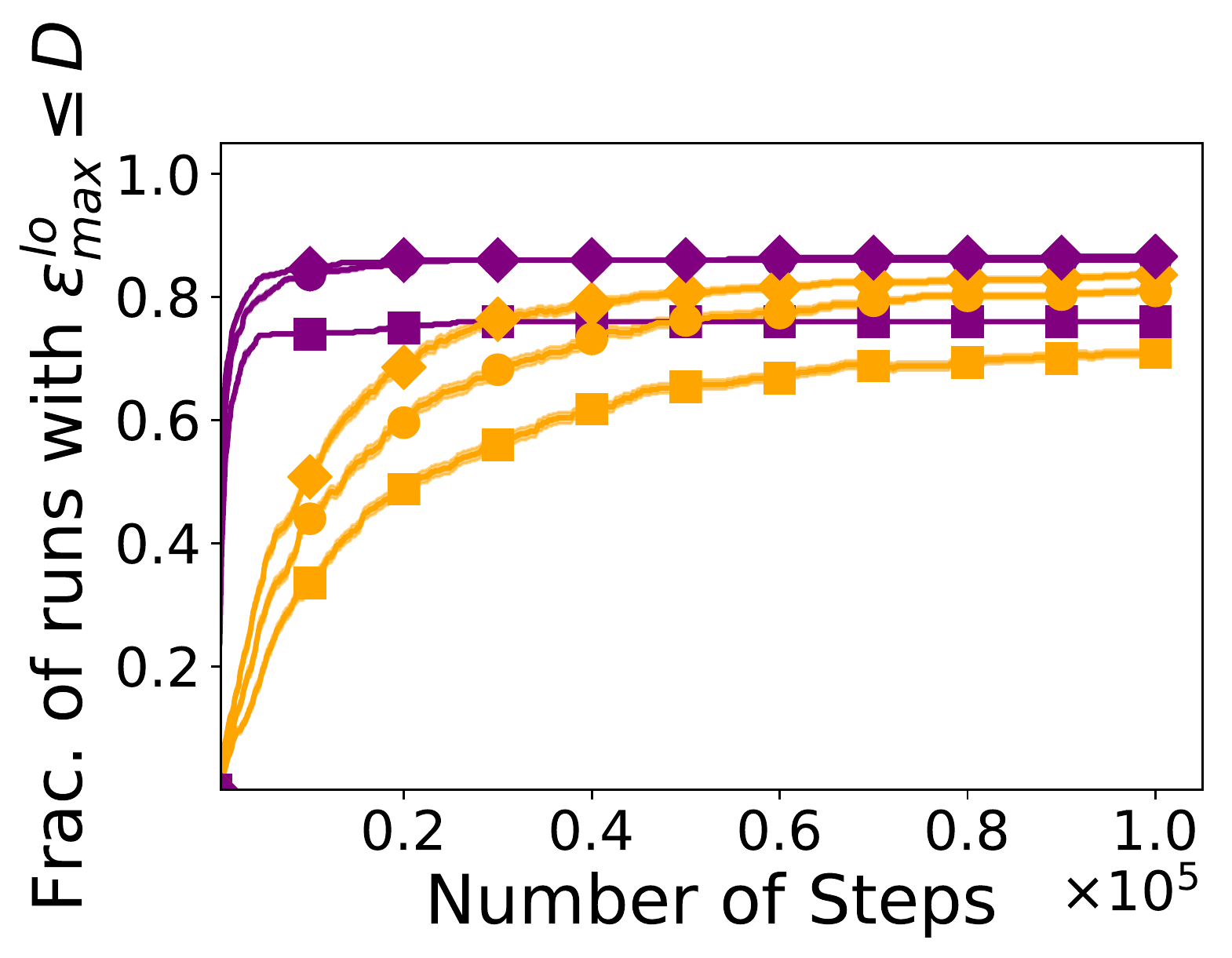}
        \captionsetup{type=figure}
        \caption{\textit{Euchre}$\boldsymbol{(12)}$ Unc.}
        \label{fig : unc_euchre_12}
    \end{subfigure}\\
    \begin{subfigure}[c]{0.24\textwidth}
        \includegraphics[width=\textwidth]{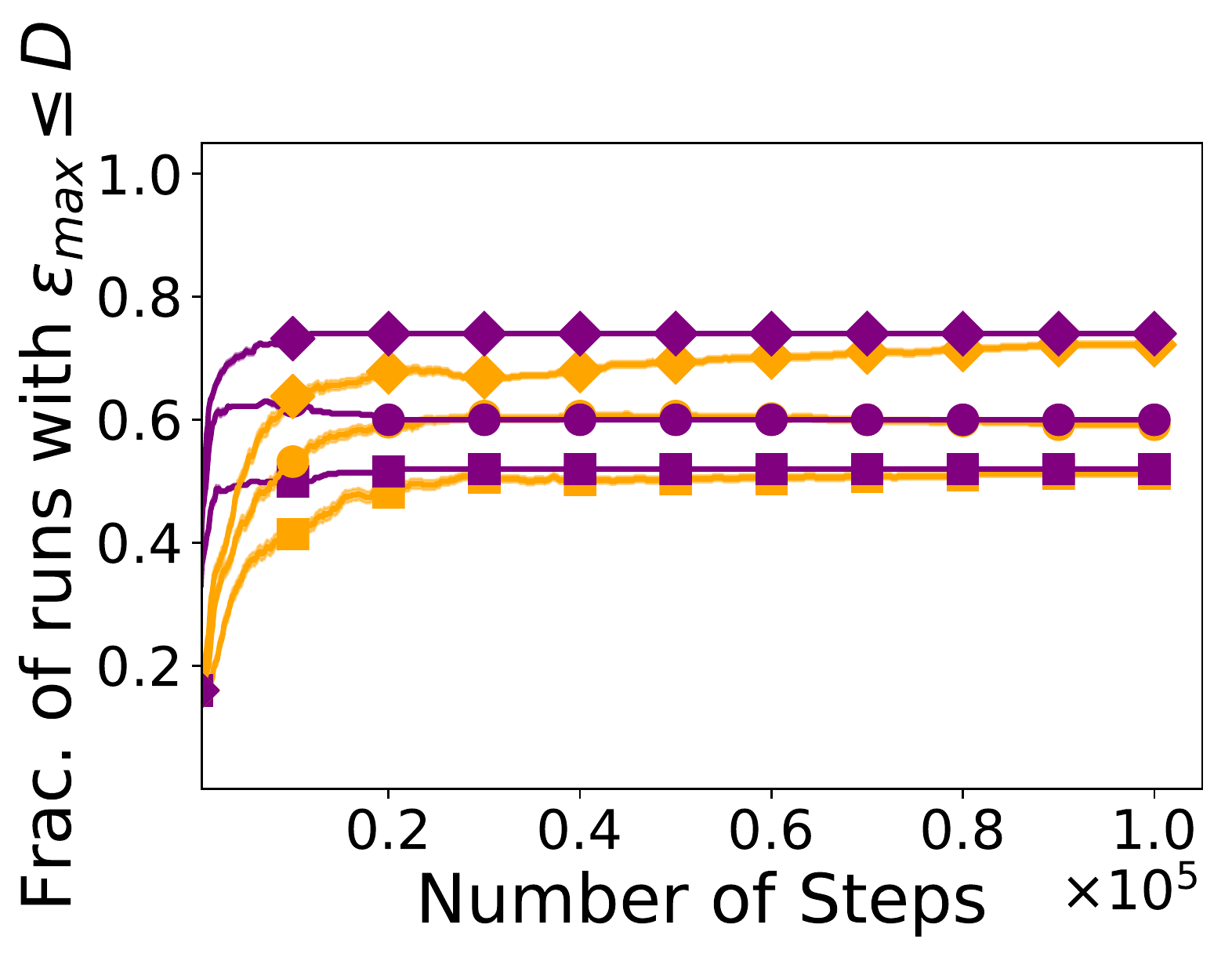}
        \captionsetup{type=figure}
        \caption{\textit{Spades}$\boldsymbol{(8)}$ Unc.}
        \label{fig : unc_spades_8}
    \end{subfigure}\hfill%
    \begin{subfigure}[c]{0.24\textwidth}
        \includegraphics[width=\textwidth]{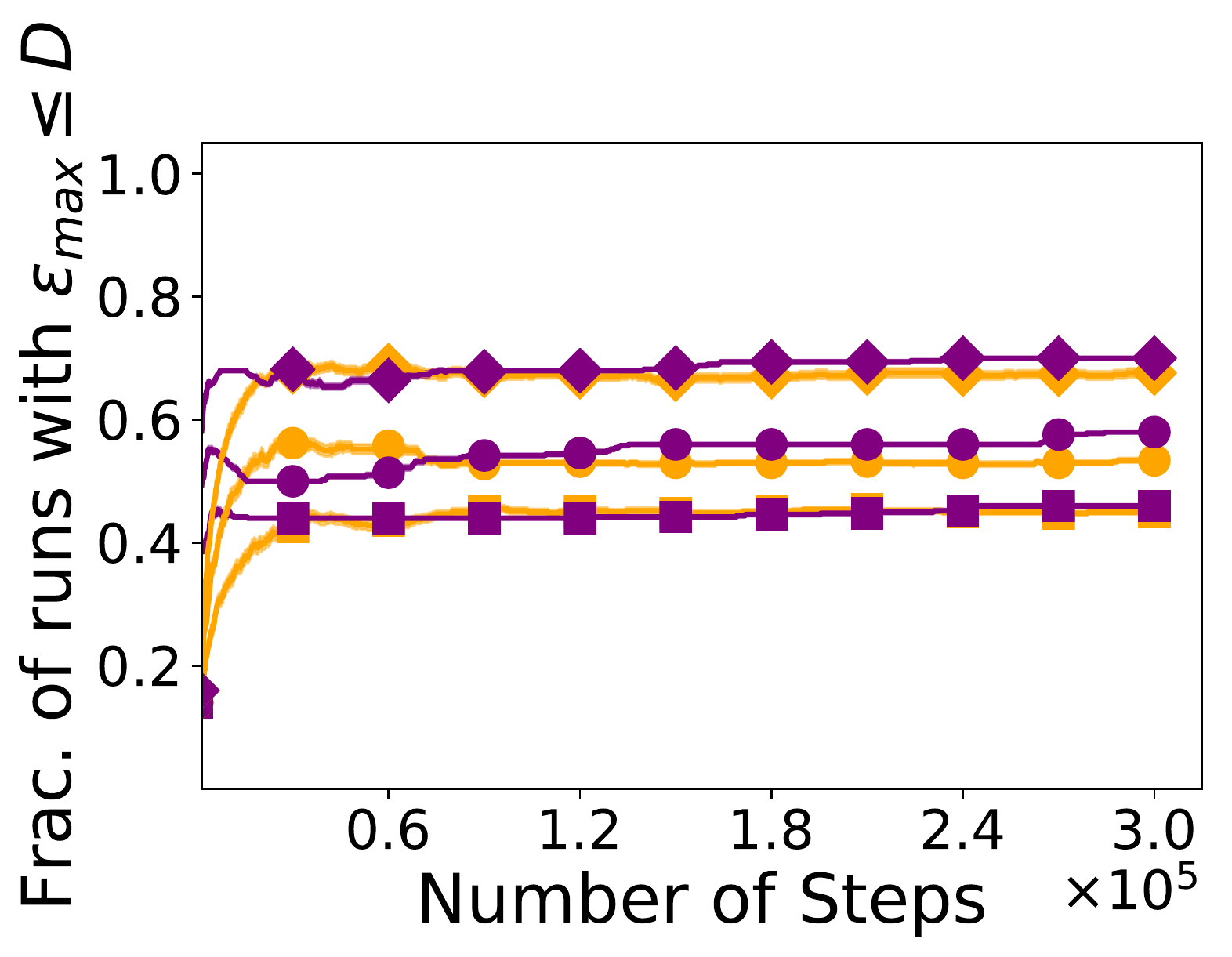}
        \captionsetup{type=figure}
        \caption{\textit{Spades}$\boldsymbol{(9)}$ Unc.}
        \label{fig : unc_spades_9}
    \end{subfigure}\hfill%
    \begin{subfigure}[c]{0.24\textwidth}
        \includegraphics[width=\textwidth]{./figures/unc/spades/env_size=10.pdf}
        \captionsetup{type=figure}
        \caption{\textit{Spades}$\boldsymbol{(10)}$ Unc.}
    \end{subfigure}\hfill%
    \begin{subfigure}[c]{0.24\textwidth}
        \includegraphics[width=\textwidth]{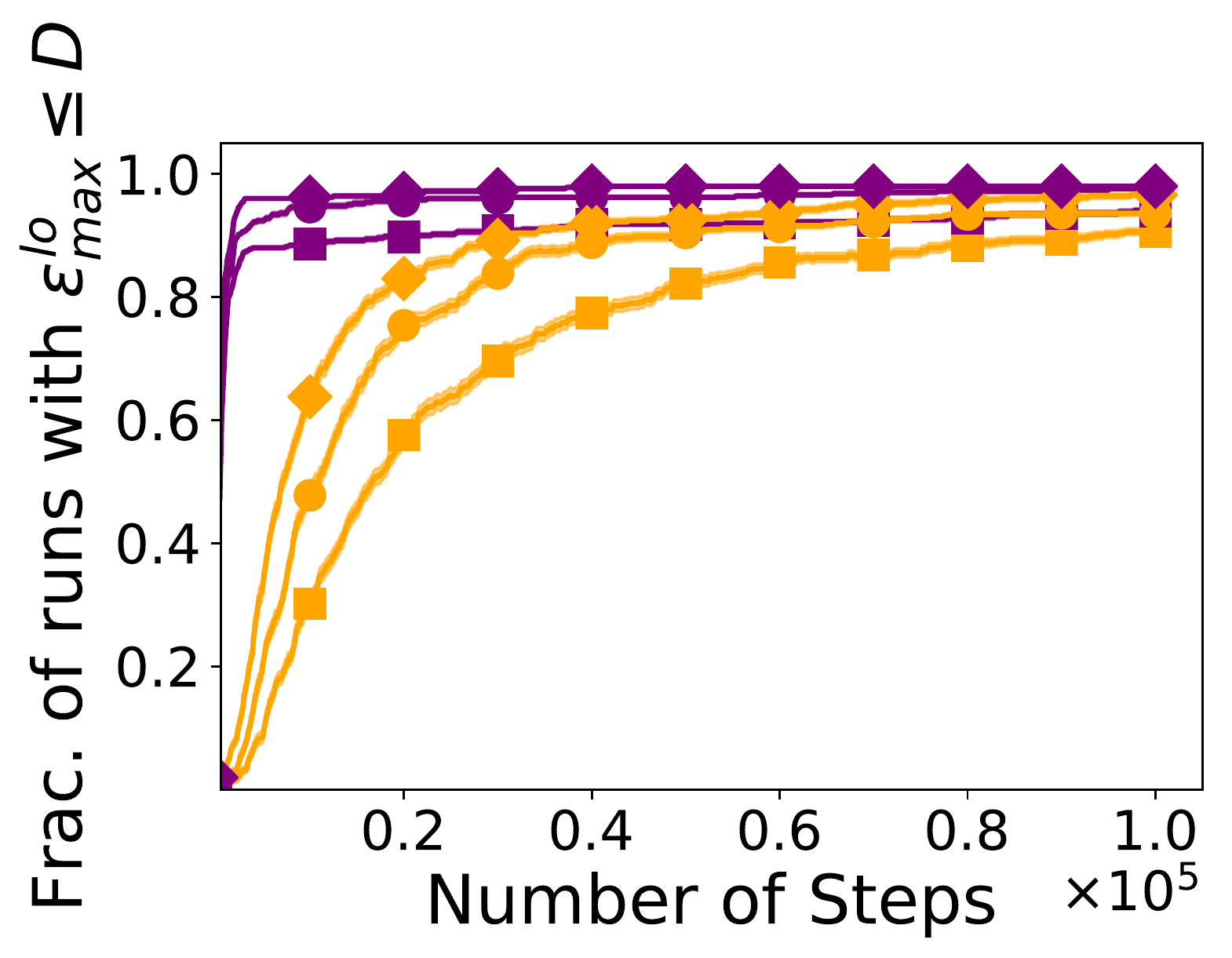}
        \captionsetup{type=figure}
        \caption{\textit{Spades}$\boldsymbol{(13)}$ Unc.}
        \label{fig : unc_spades_13}
    \end{subfigure}
\caption{Performance profiles on \textit{TeamGoofspiel}, \textit{Euchre} and \textit{Spades} with unknown context. Shaded regions show standard deviation. Each curve corresponds to a pair of method and threshold $D$.}
\label{fig: app_unc_plots}
\end{figure*}}


\end{document}